\documentclass[twoside,twocolumn]{article}

\usepackage{fitee}
\usepackage[caption=false]{subfig}
\usepackage{threeparttable}
\usepackage{multirow}
\usepackage{makecell}
\usepackage[commandnameprefix=always, final]{changes}

\usepackage[colorlinks, breaklinks = true]{hyperref}		
\addto{\captionsenglish}{%
  
}
\setaddedmarkup{\color{blue}{#1}}
\setdeletedmarkup{\color{red}{\sout{#1}}}
\newcommand{\add}[1]{\chadded{#1}}

\newcommand{\replace}[2]{\chreplaced{#2}{#1}}

\begin{document}

\title{Proposing an intelligent mesh smoothing method with graph neural networks}

\author[$\dagger$1,2]{Zhichao Wang}
\author[$\dagger$$\ddagger$1,2]{Xinhai Chen}
\author[1,2]{Junjun Yan}
\author[1,2]{Jie Liu}
\affil[1]{ Science and Technology on Parallel and Distributed Processing Laboratory, 
	\authorcr\Affilfont\it National University of Defense Technology, Changsha 410073, China}
\affil[2]{ Laboratory of Digitizing Software for Frontier Equipment, National University of Defense Technology, \authorcr\Affilfont\it  National University of Defense Technology, Changsha 410073, China}

\shortauthor{Wang et al.}	

\authmark{}



\corremailA{wangzhichao@nudt.edu.cn} 
\corremailB{chenxinhai16@nudt.edu.cn}
\emailmark{$\dagger$}	


\abstract{In CFD, mesh smoothing methods are commonly utilized to refine the mesh quality to achieve high-precision numerical simulations.  Specifically, optimization-based smoothing is used for high-quality mesh smoothing, but it incurs significant computational overhead.   Pioneer works improve its smoothing efficiency by adopting supervised learning to learn smoothing methods from high-quality meshes. However, they pose difficulty in smoothing the mesh nodes with varying degrees and also need data augmentation to address the node input sequence problem. Additionally, the required labeled high-quality meshes further limit the applicability of the proposed method. In this paper, we present GMSNet, a  lightweight neural network model for intelligent mesh smoothing.  GMSNet adopts graph neural networks to extract features of the node's neighbors and output the optimal node position. During smoothing,  we also introduce a fault-tolerance mechanism to prevent GMSNet from generating negative volume elements.   With a lightweight model, GMSNet can effectively smoothing mesh nodes with varying degrees and remain unaffected by the order of input data. A novel loss function, MetricLoss, is also developed to eliminate the need for high-quality meshes, which provides a stable and rapid convergence during training. We compare GMSNet with commonly used mesh smoothing methods on two-dimensional triangle meshes.  The experimental results show that GMSNet achieves outstanding mesh smoothing performances with 5\%  model parameters of the previous model, and attains 13.56 times faster than optimization-based smoothing.}


\doi{10.1631/FITEE.1000000}	
\code{A}
\clc{TP}


\publishyear{2018}
\vol{19}
\issue{1}
\pagestart{1}
\pageend{5}

\support{This research work was supported in part by the National Key Research and Development Program of China (2021YFB0300101), Youth Foundation of National University of Defense Technology (ZK2023-11), and National Natural Science Foundation of China (No.12102467).} 

\orcid{Zhichao Wang, https://orcid.org/0009-0007-4034-0578}	
\articleType{}

\maketitle

\section{Introduction}
\raggedbottom
With the rapid advancement of computer technology, computational fluid dynamics (CFD) has emerged as a crucial method for studying the principles of fluid dynamics. Its wide applications span diverse fields, including aerospace, hydraulic engineering, automotive engineering, and biomedicine  \citep{c1,c2,c3,c4}. Normally, CFD simulations are performed  by discretizing the governing physical equations and subsequently solving large-scale algebraic systems of discretized equations to obtain fluid variables. Discretization, a critical step in CFD, encompasses two key aspects: discretizing the governing physical equations and discretizing the computational domain \citep{c5}. The latter process, known as mesh generation, plays a fundamental role in CFD. It involves partitioning the computational domain into non-overlapping mesh elements, such as polygons in the two-dimensional region and polyhedra in the three-dimensional region \citep{c6}. The quality of the generated mesh profoundly impacts the convergence, accuracy, and efficiency of numerical simulations. Orthogonality, smoothness, distributivity and density distribution of mesh elements significantly influence the stability and convergence of the solution matrix  \citep{c7}. Consequently, the quest for high-quality mesh generation remains a vibrant and active area in CFD research. During the practical mesh generation process, the initial generated mesh often fails to meet simulation requirements. To enhance the quality of the mesh, mesh quality improvement techniques are commonly employed, including mesh smoothing \citep{c9,durand2019general,hai2021regular}, face-swapping, edge-swapping \citep{c8,c10}, point insertion/deletion \citep{escobar2005smoothing,klingner2007aggressive,guo2021adaptive}, and other techniques. Among such techniques, mesh smoothing methods are the most commonly used approach to improve mesh quality.
\begin{figure}[tbp]
    \centering
    \subfloat[]{\includegraphics[width=.43\columnwidth]{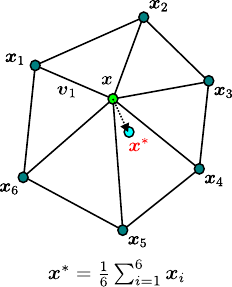}\label{fig1a}}\hspace{20pt}
    \subfloat[]{\includegraphics[width=.43\columnwidth]{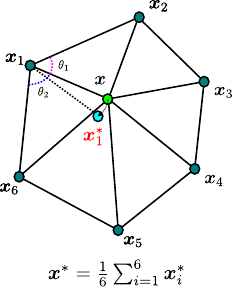}\label{fig1b}}\\
    \caption{ a) Laplacian smoothing. The \textit{StarPolygon} of point $\boldsymbol{x}$ is formed by \(\mathbf{S}(\boldsymbol{x}) = \{\boldsymbol{x}_1, \boldsymbol{x}_2, \dots, \boldsymbol{x}_6\}\) as a hexagon. The optimized mesh nodes are then located at the arithmetic average of the coordinates of the nodes in the \textit{StarPolygon}. b) Angle-Based smoothing. The figure shows the optimal position based on \(\boldsymbol{x}_1\). This is achieved by rotating $\boldsymbol{x}$ to the angle bisector of the angle where $\boldsymbol{x}_1$ is located, resulting in $\boldsymbol{x}_1^*$.  This process is repeated for other points in the mesh, and the final optimized points are obtained through arithmetic averaging. }\label{fig1}
\end{figure}

Mesh smoothing can be categorized into two main types: heuristic smoothing and optimization-based smoothing  \citep{c11}. One representative heuristic method is Laplacian smoothing  \citep{c12,pan2020hlo,sharp2020laplacian}  (shown in Figure \ref{fig1a}). In Laplacian smoothing, mesh node is placed at the arithmetic average of the coordinates of the nodes in the \textit{StarPolygon} (a polyhedron containing the node, as shown in Figure \ref{fig1}) for smoothing. This method is efficient but may produce negative volume elements  when the \textit{StarPolygon} is non-convex. Angle-based smoothing \citep{c13,guo2020angle} achieves mesh smoothing by placing mesh node on the angle bisectors of the nodes of the \textit{StarPolygon} (shown in Figure \ref{fig1b}). In addition to node-based methods, Centroidal Voronoi tessellation (CVT) smoothing  \citep{c14,du2003tetrahedral,liu2009centroidal} recalculates the Voronoi regions of each node through Lloyd iterations  \citep{c15} and relocates the point to the centroid of the Voronoi region for smoothing. To enhance the efficiency of the Lloyd algorithm, deterministic methods have been proposed for calculating the new point location  \citep{c16}. Unlike node-based methods, smoothing  methods based on element transformation  \citep{vartziotisMeshSmoothingUsing2008,vartziotisImprovedGETMeAdaptive2017,vartziotis2018getme} aim to enhance mesh quality by applying linear transformations to the mesh elements. In such methods, individual mesh elements can converge to optimal shapes through multiple iterations. Although heuristic-based mesh smoothing methods are simple and efficient, their optimization capabilities are limited. They may lead to inverted elements, and the smoothness effect heavily relies on the design of heuristic functions. On the other hand, optimization-based methods \citep{c17,canannApproachCombinedLaplacian1998,chen2004mesh,zhang2009surface} achieve mesh smoothing by optimizing the mesh quality evaluation metric in the local area.  \citet{c17} firstly formulate the mesh smoothing problem as a constrained optimization problem and utilize iterative optimization algorithms to optimize the mesh node positions. Despite the adoption of different mesh quality evaluation metrics and optimization methods in subsequent works, the optimization-based methods usually require solving  optimization problems iteratively for mesh smoothing, resulting in low efficiency.

Recently, Artificial Intelligence (AI) methods have also been widely used in mesh-related fields. Most of these research works are devoted to applying AI methods to mesh quality evaluation \citep{c27,c29,c28}, mesh density control \citep{meshingnet, meshingnet3d}, mesh generation \citep{rein,teach2mesh,mgnet}, mesh refinement \citep{bohn2021recurrent,paszynski2021deep}, and mesh adapatation \citep{tingfan2022mesh,wallwork2022e2n,fidkowski2021metric}. However, there is relatively little work on AI-based mesh smoothing. \citet{c11} firstly introduced  a supervised learning approach to imitate     optimization-based  smoothing methods with feedforward neural networks. The proposed model, NN-Smoothing, improves the efficiency of optimization-based smothing by directly giving the optimal node position. However, feedforward neural networks suffer from fixed-dimensional inputs, necessitating separate models for  mesh nodes with different degrees and data augmentation for different sequences of input nodes, thereby increasing the model's training cost.  Moreover, to train the model through supervised learning, high-quality mesh generation also incurs burdensome computational overhead.

To overcome such limitations, we present a novel mesh smoothing model, GMSNet, based on graph neural networks (GNNs)  \citep{c18} in this paper. We propose a lightweight and efficient GNN model to learn the process of mesh smoothing.  GMSNet avoids the overhead of solving the optimization problem by extracting features of the neighboring nodes of the mesh node to directly output better node position. We show that through the ability of graph neural networks to handle unstructured data, GMSNet can smooth nodes of varying degrees with a single model and elegantly solves the node input sequence problem without data augmentation. Once trained, it can be applied to smooth mesh with different shapes. We also propose a shift truncation operation to avoid introducing negative volume elements when smoothing the mesh.  Beyond the proposed GMSNet, we  introduce a novel loss function, MeticLoss, based on the mesh quality metrics to train the model, which further eliminates the overhead of generating high-quality meshes. We conduct extensive experiments among GMSNet and commonly used mesh smoothing algorithms on two-dimensional triangular meshes. The experimental results show that our model achieves a speedup of 13.56 times compared with optimization-based smoothing while achieving similar performance, and outstands all the other heuristic  smoothing algorithms. The results also indicate that GMSNet can be applied to meshes that were unseen during training.  Meanwhile, compared to previous NN-Smoothing model, GMSNet has only 5\% of its model parameter, but obtains superior mesh smoothing performance. We also validate the effectiveness of  proposed MetriLoss  with comparative experiments. We summary our contributions as follows:
\begin{enumerate}
    \item We propose a lightweight  graph neural network model, GMSNet, for intelligent mesh smoothing. GMSNet can smoothing node with varying degrees and remain unaffected by the data input order. Additionally, we offer a fault-tolerance mechanism, shift truncation, to prevent GMSNet from generating negative volume elements
    \item Basing on the mesh quality metrics, we introduce a novel loss function, MetricLoss, to train the model without necessity  for   high-quality meshes.   MetricLoss exhibits a stable and rapid convergence  during model training.
    \item  We validate the effectiveness of GMSNet \\through extensive experiments conducted on two-dimensional triangular meshes. The experimental results demonstrate that GMSNet achieves excellent   mesh smoothing performance and significantly outperforms optimi\-zation-based smoothing with a average speedup of 13.56 times. Comparative experiments are also conducted to showcase the effectiveness of the proposed MetricLoss.
\end{enumerate}

The remaining parts of this paper are organized as follows. In Section \ref{sec2}, we  introduce commonly used mesh smoothing methods and applications of neural networks in the field of mesh. In Section \ref{sec3}, we present the proposed model, providing detailed explanations of the mesh data preprocessing, design of the model's architecture, loss function and training method. In Section \ref{expr}, we conduct experiments to compare the performance of our proposed model with baseline models and discuss the effectiveness of  loss function. Finally, in the conclusion section, we summarize the entire paper and propose potential future work that can be explored.

\section{Realted work}\label{sec2}
\subsection{Heuristic mesh smoothing}
Laplacian smoothing is the most commonly used heuristic mesh smoothing method. It updates the node coordinate   to the arithmetic average of nodes in \textit{StarPolygon}. Weighted Laplacian smoothing  \citep{c19} introduces additional weights or importance factors to the neighboring nodes or edges during the smoothing process, offering more control over the smoothing effect and preserving specific features of the mesh. Smart Laplacian smoothing  \citep{field1988laplacian} includes a check before each node movement to assess whether the operation will improve the mesh quality. If the mesh quality does not improve, the movement of the mesh node is skipped, resulting in a more efficient process. Angle-based mesh smoothing achieves smoothness by considering the angles of the mesh nodes. In this method, the node is rotated to align with the angle bisectors of each node in \textit{StarPolygon}. Since the angle bisectors of the various nodes of the \textit{StarPolygon} may not coincide, the final node positions require additional calculation. This can be achieved by calculating the average of the node coordinates or by solving a least squares problem. CVT smoothing positions the node at the centroids of the Voronoi region defined by the mesh node. To enhance the efficiency of the Lloyd algorithm in the solving process, more efficient method has been designed to calculate the centroids, as described below:
\begin{equation}
    \boldsymbol{x}_i^{*}=\frac{1}{\left|\Omega_{i}\right|} \sum_{T_{j} \in \Omega_{i}}\left|T_{j}\right| C_{j}\label{cvt_equ}.
\end{equation}
where \(\boldsymbol{x}_i^{*}\) represents the new node position, \(|\Omega_{i}|\) is the total area of the node's \textit{StarPolygon}, \(|{T_{j}}|\) is the area of the $j$ th triangle, and \(C_j\) is the circumcenter of the $j$ th triangle.

It is worth noting that there is no absolute boundary between heuristic  smoothing and optimization-based  smoothing. From another perspective, heuristic mesh smoothing can also be viewed as an optimization-based approach. For instance, Laplacian smoothing can be viewed as minimizing the energy function \(E = \frac{|\mathbf{S}(\boldsymbol{x})|}{2} \sum_{i=1}^{|\mathbf{S}(\boldsymbol{x})|}\left|\boldsymbol{v}_i\right|^{2}\), where \(\boldsymbol{v}_i\) represents the edge from $\boldsymbol{x}$ to $\boldsymbol{x}_i$, and \(|\mathbf{S}(\boldsymbol{x})|\) is the number of nodes in the \textit{StarPolygon} (shown in Figure \ref{fig1a}). Similarly, Angle-based mesh smoothing can be seen as minimizing the energy function \(E = \frac{|\mathbf{S}(\boldsymbol{x})|}{2} \sum_{i=1}^{2 |\mathbf{S}(\boldsymbol{x})|} \theta_{i}^{2}\), where \(\theta_i\) is the angle between the edge from $\boldsymbol{x}$ to $\boldsymbol{x}_i$ and the edge of the polygon (shown in Figure \ref{fig1b}). The primary advantage of heuristic mesh smoothing lies in its efficiency. However, its smoothing performance often is inferior to that of optimization-based smoothing. Furthermore, the effectiveness of heuristic mesh smoothing heavily relies on   the design of the heuristic functions.

\subsection{Optimization-based  smoothing}
The mesh smoothing method based on optimization can be formalized as the following problem: Given an initial mesh with node positions \(\boldsymbol{x}_i\) and a set of constraints and the objective function $f$, the goal is to find a new  node positions \(\boldsymbol{x}_i^{*}\) that minimizes the objective function $f$. Mathematically, this can be expressed as:
\begin{equation}
    \begin{aligned}
        &\boldsymbol{x}_i^*=\underset{\boldsymbol{x}_i}{\arg \min } f\left(\boldsymbol{x}_i, \mathbf{S}\left(\boldsymbol{x}_i\right)\right),\\
        & \text{s.t.} \quad \boldsymbol{x}_i \in 	 \mathcal{X}.
    \end{aligned}
\end{equation}
where \(\boldsymbol{x}_i^*\) represents the new position of the mesh node \(\boldsymbol{x}_i\), \(f\) is some mesh quality evaluation function, \(\mathcal{X}\) is the feasible set satisfying constraints on \(\boldsymbol{x}_i^*\), and \(\mathbf{S}(\boldsymbol{x}_i)\) is the set of nodes of the \textit{StarPolygon} of \(\boldsymbol{x}_i\). It is important to mention that the input of this function also includes the connectivity between nodes, which is omitted here for clarity. Different choices of evaluation functions lead to different mesh smoothing methods, reflecting our emphasis on different mesh qualities. Commonly used evaluation functions include the maximum minimum angle \citep{c22}, aspect ratio, and distort ratio, among others  \citep{c17}. If the function \(f\) is differentiable, the optimal point of this constrained optimization problem may be solved by setting the gradient \(\nabla f=0\) to obtain an explicit expression. However, in most cases, it is difficult to derive explicit expressions for \(\boldsymbol{x}_i^*\) as in Laplacian smoothing and Angle-based smoothing. Iterative methods are often used to solve \(\boldsymbol{x}_i^*\), which are of low efficiency. Therefore, developing an efficient way to solve the optimal positions is a problem that needs to be addressed.

\subsection{Neural networks in mesh-related fields}
Inspired by the neurons in the human brain, artificial neural networks are utilized to learn complex function mappings. They find extensive applications in machine learning and artificial intelligence, addressing tasks like image recognition, speech recognition, natural language processing, and more \citep{c23,c24,c25,c26}. 
Recently, neural networks have found significant application in various domains related to meshes. In the realm of mesh quality evaluation, GridNet, a convolutional neural network  model, was introduced by \citet{c27}, along with the NACA-Market dataset, to facilitate automated evaluation of structured mesh quality. Extending this notion to unstructured meshes, \citet{c28} employed graph neural networks for mesh quality assessment. In the pursuit of refining mesh distribution, \citet{meshingnet} employed an artificial neural network to enhance conventional mesh generation software, enabling prediction of local mesh density throughout the domain. This approach was further expanded to  tetrahedral meshes, as demonstrated convincingly through extensive testing \citep{meshingnet3d}.    In the domain of intelligent mesh generation, \citet{rein} presented an algorithm that leverages global structural information from point clouds to achieve high-quality mesh reconstruction. Similarly, \citet{teach2mesh} harnessed data extracted from meshed contours to train neural networks, enabling accurate approximation of the number, placement, and inter\-connectivity of nodes within the meshing domain. Taking a novel differential approach, \citet{mgnet} introduced MGNet, an unsupervised neural network methodology for generating structured meshes, yielding promising results. Beyond generation, artificial intelligence techniques have also made significant contributions to mesh refinement and adaptation. \citet{bohn2021recurrent} employed recurrent neural networks to learn optimal mesh refinement algorithms, establishing its prowess as an effective black-box tool for enhancing a wide spectrum of partial differential equations. Enriching variational mesh adaptation, \citet{tingfan2022mesh} seamlessly integrated a machine learning regression model to expedite flow field estimation on updated meshes. Meanwhile, \citet{wallwork2022e2n} devised a data-driven goal-oriented mesh adaptation strategy, underpinned by a trained neural network, effectively \\supplanting the computationally expensive error estimation phase in the adaptation process. Furthermore, \citet{fidkowski2021metric} ingeniously employed Artificial Neural Networks to ascertain optimal aniso\-tropy in computational meshes, yielding enhanced mesh efficiency in comparison to conventional methods.  In term of mesh smoothing, NN-Smoothing \citep{c11}  imitates optimization-based mesh smoothing  using feedforward neural networks, significantly enhancing the efficiency of optimization-based mesh smoothing. However, separate models training and expensive high-quality mesh generation incur significant computational overhead.

In addition to conventional deep learning algorithms, graph neural networks   \citep{c18} were introduced to enhance the learning capacity of artificial intelligence methods for handling non-structured data. GNNs utilize graph convolutions  \citep{c30} to incorporate the topological connections in the feature learning process on graphs. As meshes can be naturally represented as graph data, GNNs have found extensive applications in various computational fluid dynamics fields, including mesh refinement, flow field simulation, turbulence modeling, and more \citep{c31,c32,c33,c34,c35}. Therefore, we argue that applying GNNs to mesh smoothing is a promising and effective solution.
\begin{figure*}[htb]
    \centering
    \includegraphics[width=\linewidth]{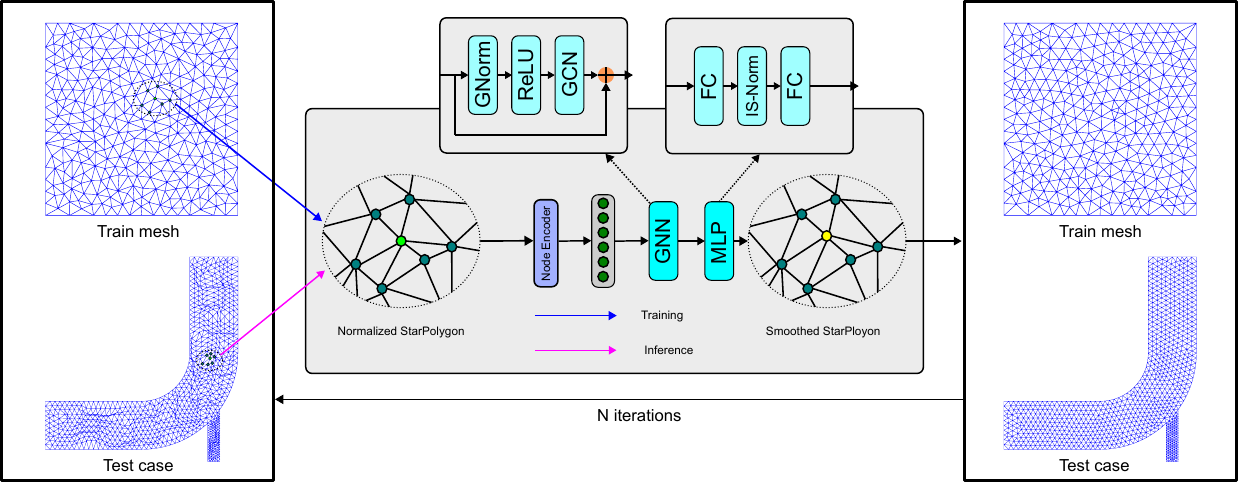}
    \caption{The architecture of GMSNet. GMSNet smoothes the mesh by calculating the optimized position for each mesh node. The process begins with the normalization of the input \textit{StarPolygon}. Next, feature transformation is performed on the normalized features, and information from the \textit{StarPolygon} nodes is integrated using graph convolution. Finally, the model predicts the optimized node positions through  fully connected (FC) layers. In the figure, \textbf{GNorm} represents the GraphNorm operation, and \textbf{IS-Norm} represents the InstanceNorm operation. The model iterates $N$ times over the entire mesh to perform the smoothing operation. After training, GMSNet can be applied to previously unseen meshes, such as the pipe's mesh.}\label{fig_model}
\end{figure*}

\section{Methodology}\label{sec3}
\subsection{Problem formulation}
The mesh smoothing problem involves enhancing the quality of a mesh by adjusting the positions of its nodes while maintaining the connectivity between them. Mesh smoothing processes are defined as functions that operate on the mesh, taking the mesh nodes and their connections as inputs and providing new coordinates for the mesh nodes as outputs. For each mesh node, the input of the smoothing function is itself and its \textit{StarPolygon}, which comprises the node's one-ring neighbors. In this paper, we define the mesh as a node graph. Specifically, given a node $\boldsymbol{x}_0$ and its \textit{StarPolygon}, we represent it with a graph \(\mathcal{G}=(\mathcal{V},\mathcal{E})\), where \(\mathcal{V}=\{\boldsymbol{x}_0,\boldsymbol{x}_{1},\boldsymbol{x}_2,\dots,\boldsymbol{x}_{n}\}\) represents the node and nodes in the \textit{StarPolygon}, \( \boldsymbol x_i \) is the coordinate of node \( i \) in \textit{StarPolygon}, \(n\) is the number of nodes in the \textit{StarPolygon}, and \(\mathcal{E}=\{(i,j) \mid \text{if } \boldsymbol{x}_i \text{ is connected with } \boldsymbol{x}_j\}\) represents the connections between the nodes.
A typical smoothing process is iterative, where at the \(t\) th iteration step, the initial node graph is denoted as \(\mathcal{G}_t\), and the optimized node position for the center node can be represented as:
\begin{equation}
    \boldsymbol{x}_0^{t+1}=\mathcal{F}(\mathcal{G}_t)=\mathcal{F}(\mathcal{V}_t,\mathcal{E}).
\end{equation}
In this paper, we employ graph neural networks to learn the function \(\mathcal{F}\) for mesh smoothing.
\begin{figure}[tb]
    \centering
    \includegraphics[width=\columnwidth]{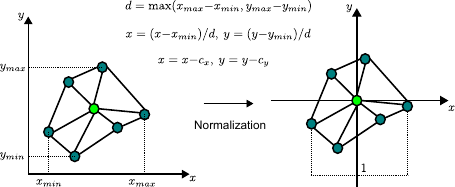}
    \caption{Data normalization. The \textit{StarPolygon} to be processed is normalized to the center of the coordinate axis. In the figure, \(d\) represents the maximum size of the \textit{StarPolygon}, \(x\) and \(y\) represent the coordinates to be scaled, and \(c_x\) and \(c_y\) are the normalized coordinates of the target node after the scaling.}\label{norm}
\end{figure}

\subsection{GMSNet}
\subsubsection{Lightweight model design}

The mesh smoothing model based on graph neural networks is depicted in Figure \ref{fig_model}. Given a graph consisting of mesh nodes and edges, our goal is to compute a more optimal mesh node position for each mesh node to smoothing the mesh. 
In the model, we use the coordinate positions as node features, denoted as \(\mathbf{X} \in \mathbb{R}^{(n+1) \times 2}\) ($n$ nodes in the \textit{StarPolygon} and one free node to move). The connectivity between nodes is represented by the adjacency matrix \(\mathbf{A} \in \mathbb{R}^{(n+1) \times (n+1)}\) (node index \( i \) is omitted for clarity). 

To ensure the model's scale invariance, we apply node normalization during processing. The model normalizes the node input using min-max normalization on \(\mathbf{X}\), restricting it to the range of 0 and 1, and subsequently performs a translation to center the node around the coordinate origin. After model processing, we employ an affine transformation to map the node positions back to their original scale. Data normalization is illustrated in Figure \ref{norm}. After normalizing the data, we transform the features by a linear layer, which can be expressed as
\begin{equation}
    \mathbf{X}_h=\mathrm{Norm}(\mathbf{X})\mathbf{W}_l +\mathbf{b}_l.
\end{equation}
where \( \mathrm{Norm} \) represents the normalization operation, \(\mathbf{W}_l \in \mathbb{R}^{2 \times H}\) and $\mathbf{b}_l$ is the transformation matrix and bias of the linear layer, \( H \) is the dimension of the hidden feature, and \(\mathbf{X}_h \in \mathbb{R}^{(n+1) \times H} \) is the output of the linear layer.
Subsequently, we employ a residual graph convolutional network (GCN) layer \citep{c30,c36} to compute the features of the hidden layer based on the features of the nodes in the \textit{StarPolygon}. The process involves normalizing the features outputted by the linear layer using GraphNorm  \citep{c37}, followed by an activation layer, a convolutional layer, and a summation layer to calculate the hidden features of the nodes. This process can be represented as:
\begin{align}
    \mathbf{\hat{X}}_h&=\mathrm{GraphNorm}(\mathbf{X}_h),  \\
    \mathbf{X}_g&=\mathrm{GCN}(\mathrm{ReLU}(\mathbf{\hat{X}}_h),\mathbf{\tilde{A}})+\mathbf{\hat{X}}_h\\
    &=\mathbf{\tilde{A}}[\mathrm{ReLU}(\mathbf{\hat{X}}_h)]\mathbf{W}_g+\mathbf{\hat{X}}_h.
\end{align}
where \( \mathbf{\tilde{A} }\) is the normalized adjacency matrix\footnote{$\mathbf{\hat{A}}=\mathbf{A}+\mathbf{I}$,$\mathbf{\hat{D}}$ is the degree matrix of $\mathbf{\hat{A}}$, and \( \mathbf{\tilde{A}}=\mathbf{\hat{D}}^{-\frac{1}{2}}\mathbf{\hat{A}}\mathbf{\hat{D}}^{-\frac{1}{2}} \) }, ReLU is activation function, $\mathbf{W}_g$ is the parameter of GCN layer, and $\mathbf{X}_g \in \mathbb{R}^{(n+1) \times H}$ represents the final features output by the residual GCN layer. The final position of the free node is obtained through a two-layer fully connected neural network with InstanceNorm  \citep{c38}. We only need to change the  position of the free node , so we  extract the feature of the free node and decode it through an MLP. Let $i_c$ be the index of the free node, then the optimized node position is given by:
\begin{equation}
    \boldsymbol{x}^*=\mathrm{MLP}({\mathbf{X}_g}_{[i_c,:]}).
\end{equation}

The smoothing algorithm should possess the capability to handle nodes with varying degrees and remain unaffected by the order of node inputs. In the NN-Smoothing model, handling nodes with different degrees is achieved by training separate models, while the order of node inputs is addressed through data augmentation by varying the starting node in the ring of \textit{StarPolygon}. However, owing to the permutation invariance property of GNNs, where the output remains unaffected by permutations in the input order (e.g., in the sum or mean function), GMSNet  remains unaffected by changes in the node input sequence. Additionally, it can also smoothing the nodes with varying degrees, as there are no restrictions on the number of neighbors a node can have in GNNs.
\begin{table*}[htbp]
    \centering
    \footnotesize
    \caption{Comparison among optimization-based smoothing, NN-Smoothing and GMSNet}
    \label{tab:model_comp}
    \begin{threeparttable}
        \begin{tabular}{ccccc}
            \toprule
            \thead{Method} & {Speed}  &  {Labeled high-quality mesh}  & {Varying node degrees} & {Node input order} \\
            \midrule
            OptimSmoothing\tnote{1} & Slow  & Not acquiring & Not affected & Not affected \\
            NN-Smoothing & Fast  & Acquiring & Training separate models & Performing data augmentation \\
            GMSNet & Fast  & Not acquiring & Not affected & Not affected \\
            \bottomrule
        \end{tabular}
        \begin{tablenotes}
            \item[1] Optimization-based smoothing.
        \end{tablenotes}
    \end{threeparttable}
\end{table*}

\subsubsection{Model  training}
In the NN-Smoothing method, the model is trained using supervised learning, and the labeled high-quality meshes are generated using optimization-based smoothing, which is a time-consuming task. In contrast, the proposed GMSNet  directly learns the optimization process for mesh smoothing. This process can be illustrated by comparing it with optimization-based smoothing.
In optimization-based smoothing, taking gradient descent as the optimization method and given a mesh quality function \(f\) to be optimized, the objective is to find its minimum value. In the \(k\) th iteration, the optimization algorithm updates the position as \(\boldsymbol{x}_{k+1} = \boldsymbol{x}_k - \alpha \nabla_{\boldsymbol{x}} f(\boldsymbol{x}_k,\mathbf{S}(\boldsymbol{x}_0))\), where \(\nabla_{\boldsymbol{x}} f(\boldsymbol{x}_k,\mathbf{S}(\boldsymbol{x}_0))\) is the gradient of \(f\) and \(\alpha\) is the step size. By iteratively updating \(\boldsymbol{x}_k\), the function converges to a local or global optimum \(\boldsymbol{x}^*\).
In contrast, NN-Smoothing directly predicts the optimal point for mesh smoothing, i.e., \(\boldsymbol{\hat{x}}^* = \mathrm{NN}(\boldsymbol{x}_0,\mathbf{S}(\boldsymbol{x}_0))\). However, this approach requires labeled high-quality meshes generated through the optimization algorithm, which is time-consuming.
On the other hand, the GMSNet model does not require labeled data to learn the smoothing process for mesh nodes. Instead, the training process is driven by the quality evaluation metric of the mesh elements, which can be expressed as:
\begin{equation}
    \begin{aligned}
        \mathbf W^*&={\arg\min}_{\mathbf{W}}\mathcal{L}(\boldsymbol{\hat{x}}^*,\mathbf{S}(\boldsymbol{x}_0))\\&={\arg\min}_{\mathbf{W}}\mathcal{L}(\mathcal{F}_{\mathbf W}(\boldsymbol{x}_0,\mathbf{S}(\boldsymbol{x}_0)),\mathbf{S}(\boldsymbol{x}_0)). 
    \end{aligned}
\end{equation}
where $\mathcal{L}$ is the loss function  constructed using a mesh quality metric, $\mathcal{F}_{\mathbf W}$ is learned through the proposed model, and ${\mathbf W}$ is the parameters of the model.
The loss function is based on the mesh element evaluation metric, and the model optimizes the positions of mesh nodes by minimizing this function.
\begin{figure}[tb]
    \centering
    \includegraphics[width=\columnwidth]{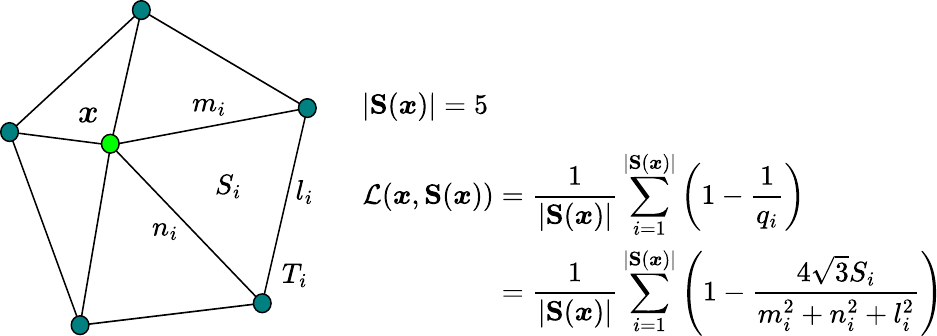}
    \caption{MetricLoss.   We use the mean transformed aspect ratio as the loss function for model optimization.}\label{metricloss}
\end{figure}

The main distinction between this approach and NN-Smoothing lies in the training data. In the proposed method, there are no high-quality meshes provided for the labels. Instead, the learning process relies solely on minimizing the mesh quality metric function. The primary divergence from optimization-based mesh smoothing methods is that this approach directly offers optimized positions for mesh nodes without the need to solve an optimization problem, resulting in a significant improvement in the efficiency of mesh smoothing. A comprehensive comparison of the three aforementioned methods is shown in Table \ref{tab:model_comp}.

\subsubsection{MetricLoss }\label{sec:loss_design}

In this section, we introduce a loss function based on mesh quality  metrics, called MetricLoss, as the optimization objective function for the model. By minimizing this loss function, the model can learn how to smooth the  mesh to improve its quality. There are several metrics available to evaluate the quality of mesh elements, including the maximum angle, minimum angle, Jacobian matrix, aspect ratio, and others. In our case, we adopt the aspect ratio \(q = \frac{{m^2 + n^2 + l^2}}{{4 \sqrt{3}  S}}\) to assess the mesh quality, where \(m\), \(n\), \(l\) are the edges of the triangle, and \(S\) is the area of the triangle. For an equilateral triangle, this value is 1, while for a degenerate triangle, it approaches \(+\infty\). However, the range of this metric is too large, which can cause gradient explosion, especially for poor-shape elements. To address this issue, we normalize  this metric its domain to  \( [0,1] \)  through function $f(q)=1-\frac{1}{q}$. For equilateral triangles, this transformed metric is 0, while for degenerate triangles, it is 1. The MetricLoss is designed as the mean value of the transformed metric of elements in its \textit{StarPloygon}, which can be expressed as:
\begin{equation}
    \begin{aligned}
           \mathcal{L}(\boldsymbol{x},\mathbf{S}(\boldsymbol{x}))&=\frac{1}{|\mathbf{S}(\boldsymbol{x})|}\sum_{i=1}^{|\mathbf{S}(\boldsymbol{x})|}{(1-\frac{1}{q_i})}\\&=\frac{1}{|\mathbf{S}(\boldsymbol{x})|}\sum_{i=1}^{|\mathbf{S}(\boldsymbol{x})|}{(1-\frac{4\sqrt{3} S_i}{m_i^2+
                n_i^2+l_i^2} )}.  
    \end{aligned}
\end{equation}
where $|\mathbf{S}(\boldsymbol{x})|$ is the number of nodes in the \textit{StarPolygon}, \(m_i\),\(n_i\),\(l_i\) are the edges of triangle $T_i$, and $q_i$ is the aspect ratio of $T_i$ (see Figure \ref{metricloss} for details). We validated the effectiveness of our designed loss function in {Section \ref{sec:loss_func}}. 

\subsubsection{Shift truncation}
\begin{figure}[tb]
    \centering
    \subfloat[]{\includegraphics[height=0.45\columnwidth]{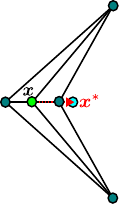}\label{LaplaceOut}}\hspace{20pt}
    \subfloat[]{\includegraphics[height=0.45\columnwidth]{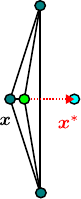}\label{CVTOut}}\hspace{20pt}
    \subfloat[]{\includegraphics[height=0.45\columnwidth]{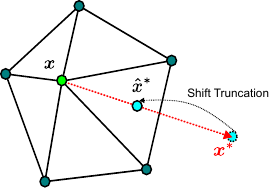}\label{Trunc}}\\
    \caption{  a) For non-convex \textit{StarPolygons}, Laplace smoothing may  move nodes outside the \textit{StarPolygon}.		b) Computing the circumcenter of triangles is necessary for CVT smoothing. For highly distorted mesh element, it circumcenter is far away from the \textit{StarPolygon}, resulting in negative volume elements.  b) Shift truncation. In the training and inference phases of the model, we truncate the shift to avoid generating negative volume elements. }\label{shift_trunc}
\end{figure}

In the process of mesh generation and mesh smoothing, it is   important to ensure that the generated mesh avoids negative volume elements. However, mesh smoothing algorithms can sometimes lead to the generation of negative volume elements, as depicted in Figure \ref{shift_trunc}. For instance, in the case of Laplacian smoothing, when dealing with non-convex \textit{StarPolygons}, negative volume elements may arise (as illustrated in Figure \ref{LaplaceOut}). Similarly, during CVT smoothing, the calculation of the Voronoi centroid for elongated mesh elements can result in shifts that extend far away from the \textit{StarPolygon} region, as shown in Figure \ref{CVTOut} (in the case of degenerate mesh elements, the centroid position may even be at an infinite distance). Additionally, in optimization algorithms, using a uniform step size for different scales of mesh elements can also lead to the production of negative volume elements, as the optimization process may overshoot the optimal position.
\begin{figure}[tb]
    \centering
    \subfloat[Train mesh after  training model for 1 epoch]{\includegraphics[width=.35\columnwidth]{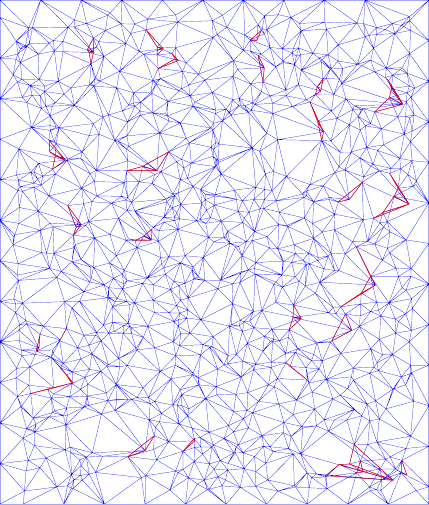}\label{fig:shift_train1}}\qquad
    \subfloat[Train mesh  after  training model for 10 epochs]{\includegraphics[width=.55\columnwidth]{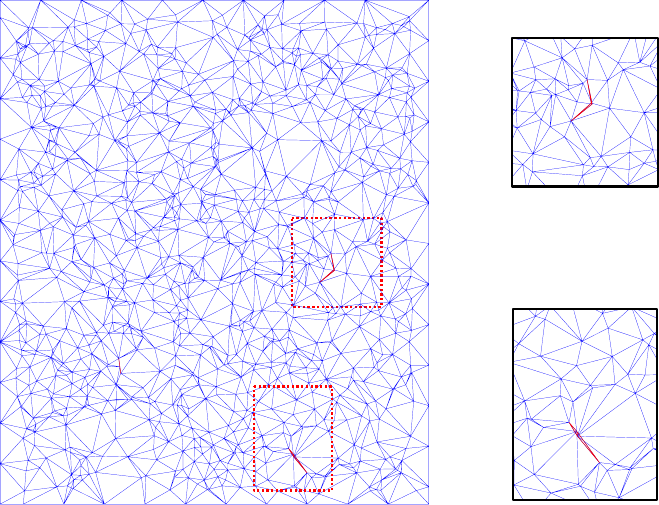}\label{fig:shift_train10}}\\
    \subfloat[Test mesh  after  training model for 1 epoch]{\includegraphics[width=.37\columnwidth]{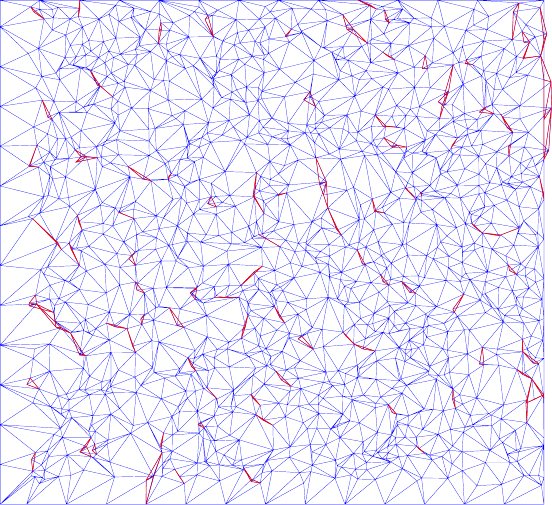}\label{fig:shift_val1}}\qquad
    \subfloat[Test mesh  after  training model for 10  epochs]{\includegraphics[width=.53\columnwidth]{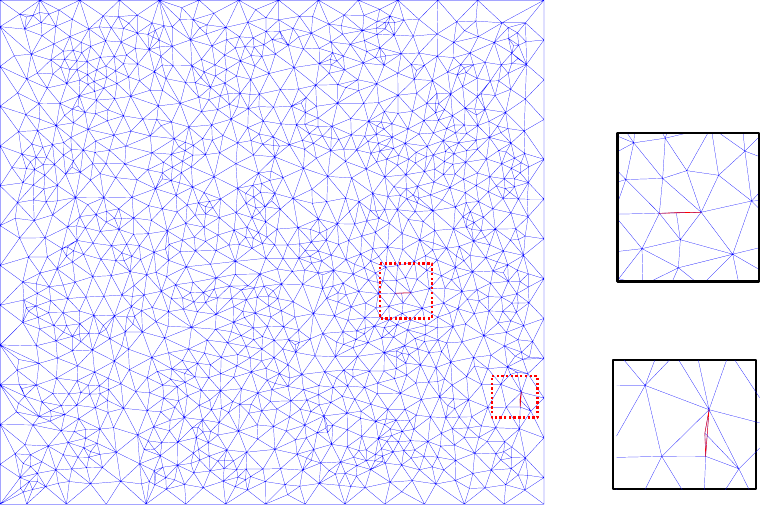}\label{fig:shift_val10}}\\
    \caption{The essential of shift truncation. The red elements in the figure represent negative volume elements. }\label{fig:shift_trunc}
\end{figure}
\begin{algorithm}[tb]\small
    \centering
    \caption{Shift truncation operation} \label{trun_algo}
    \begin{algorithmic}[1]
        \STATE {Mesh Node $i$ original position $\boldsymbol{x}_i$, 	shift $\Delta \boldsymbol{x}_i$, optimized position  $\boldsymbol{x}_i^*=\boldsymbol{x}_i+\Delta \boldsymbol{x}_i$}
        
        \WHILE{ $\boldsymbol{x}_i^*$ results in negative elements}
        \STATE{$\Delta \boldsymbol{x}_i=0.5\Delta \boldsymbol{x}_i$}
        \STATE{$\boldsymbol{x}_i^*=\boldsymbol{x}_i+\Delta \boldsymbol{x}_i$}
        \ENDWHILE
    \end{algorithmic}
\end{algorithm}	
\begin{algorithm}[tb]\small
    \centering
    \caption{Training of GMSNet} \label{train_alg}
    \begin{algorithmic}[1]
        \STATE {Mesh dataset $\mathbf{M}=\{{\mathcal{M}_i}\}_{i=1}^n$, MetricLoss $\mathcal{L}$}
        \STATE {Training epochs $N$, batch size $\mathcal{B}$, learning rate $\alpha$}
        \STATE {GMSNet parameters $\mathbf{W}$}
        
        \FOR{ $j \gets  1 ~\mathrm{to}~ N $ }
        \FOR{mesh $\mathcal{M}_i$ in $\mathbf{M}$}
        \STATE{Sample $\mathcal{B}$ mesh nodes from $\mathcal{M}_i$: \( \{\boldsymbol{x}_i\}_{i=1}^{\mathcal{B}} \)}
        \STATE{Compute the optimized node position: $\boldsymbol{x}_i^*=\mathrm{GMSNet}(\boldsymbol{x}_i, \mathbf{S}(\boldsymbol{x}_i))$ for $\boldsymbol{x}_i
            \in \{\boldsymbol{x}_i\}_{i=1}^{\mathcal{B}}$}
        \STATE{Shift truncate $\Delta \boldsymbol{x}_i$ if $\Delta \boldsymbol{x}_i$ results in negative volume elements}
        \ENDFOR
        \STATE{Update model parameters: $\mathbf{W} \gets $ $\mathbf{W}-\frac{\alpha}{\mathcal{B}}\sum_{i=1}^{\mathcal{B}}\nabla\mathcal{L}(\boldsymbol{x}_i^*, \mathbf{S}(\boldsymbol{x}_i^*))$}
        \algocomment{Here we use stochastic gradient descent as the optimization method.}
        \ENDFOR
    \end{algorithmic}
\end{algorithm}	

Negative volume elements can also occur during the training and inference stage of neural network-based smoothing algorithms.  To show this, we visualized the output mesh after the first epoch of training. It can be observed that a large portion of the node updates resulted in negative volume elements, as shown in Figure \ref{fig:shift_train1}. However, as the model continues training, it learns the smoothing process, and the number of negative volume elements decreases, as depicted in Figure \ref{fig:shift_train10}. We also study the impact of  shift truncation in the model predictions. The experimental results  are shown in Figure \ref{fig:shift_val1} and \ref{fig:shift_val1}.  It is evident that for certain highly distorted elements, the model may produce negative volume elements despite being trained for  epochs.  The generation of negative volume elements does not disrupt the learning process. With the model training, the occurrence of negative volume elements gradually decreases. However, due to the uncertainty of neural networks, although rare, it is still possible for the model to introduce negative volume elements during the inference stage.  Hence, a method is required to prevent the occurrence of negative volume elements. The simplest approach is to set the displacements that result in negative volume elements to zero. However, this approach hinders the update of poorly shaped mesh elements, which is precisely the optimization goal of mesh smoothing. Therefore, we have adopted a line search method to handle negative volume elements, as depicted in Figure \ref{Trunc} and Algorithm \ref{trun_algo}. We repeatedly half the shift which introduces the negative volume elements until no negative volume element is generated. With the aforementioned method, the training of GMSNet is described in Algorithm \ref{train_alg}.

\section{Experiments}\label{expr}
\subsection{Exprimental Setup}

In this section, we conducted a comprehensive performance comparison of the proposed GMSNet  with five baseline  smoothing methods (algorithms). The baseline methods we evaluated are as follows:  Laplacian smoothing \citep{field1988laplacian}, Angle-based smoothing \citep{c13}, CVT smoothing \citep{c14}, GETMe smoothing \citep{vartziotis2018getme}, Optimization-based smoothing, and NN-Smoothing \citep{c11}. Below are the implementation details for each baseline model:
\begin{itemize}
    \item All models adopted the \textbf{sequential updating}, where the mesh nodes are directly updated after each optimization step (as opposed to computing the updates for all nodes and then updating them together).
    \item \add{All models are implemented entirely in Python, with their sole distinction lying in how they update the positions of mesh nodes.}
    \item Smart Laplacian smoothing was adopted to prevent negative volume elements.
    \item In the Angle-based smoothing, the final node positions were obtained by averaging the optimized positions computed for each angle in the \textit{StarPolygon}.
    \item The CVT smoothing employed  the Equation \ref{cvt_equ} to improve the algorithm's performance.
    \item The element transformation-based smoothing methods, GETMeSmoothing, is implemented in accordance with the algorithms outlined in   \textit{The GETMe Mesh  Smoothing Framework} \citep{vartziotis2018getme}. The parameter configuration utilized in this implementation includes $\theta=\frac{\pi}{3}$, $\lambda=0$, and $\varrho=1$. To ensure a fair comparison, a sequential version was employed. 
    \item The Optimization-based smoothing used MetricLoss defined in section \ref{sec:loss_design}  as the objective function and employed Adam \citep{
        c40} as the optimizer. Each node was optimized for a maximum of 20 iterations.
    \item The NN-Smoothing  followed the implementation approach in the original paper. For nodes with different degrees, we trained different models to smooth the meshes. Additionally, Laplacian smoothing was used to handle nodes with degrees other than 3, 4, 5, 6, 7, 8, and 9.
\end{itemize}

We trained the model solely using two-dimensional triangle meshes, comprising a total of 20  meshes with a square domain. The dataset was split into training, validation, and testing sets in a ratio of 6:2:2. Each mesh has distinct sizes and densities, randomly generated before the training. The mesh nodes are positioned randomly within the geometric domain, and the meshes are generated using Delaunay triangulation  \citep{c39}.  Figures of  mesh examples can be found in \textbf{Appendix \ref{appendix_data}}.

In the experiment, we utilized the final optimization results obtained from the Optimization-based smoothing as the training labels for the NN-Smoothing model. Simultaneously, we trained the GMSNet  without incorporating labels during the training process using the same dataset. For both neural network models, we employed Adam \citep{c40} as the optimizer with an initial learning rate of 1e-2. Throughout the training process, the learning rate was dynamically adjusted based on the performance on the validation set. In each training epoch, we randomly sampled 32 mesh nodes from each mesh to train the models.  Despite training with only a partial sampling of mesh nodes, the models   converge effectively. \add{NN-Smoothing and GMSNet models were trained on an NVIDIA Telsa P100 GPU. During the testing of model performance, for fair comparison, all the models were run on the  Intel Core i7-12700H CPU.}

\subsection{Experimental results}

\begin{figure*}[tb]
    \centering
    \subfloat{\includegraphics[width=0.5\columnwidth]{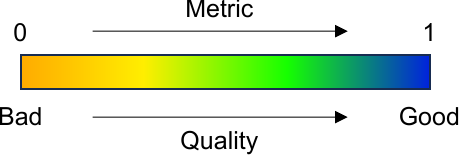}}\\ \setcounter{subfigure}{0}
    \subfloat[Square mesh before smoothing]{\includegraphics[width=0.45\columnwidth]{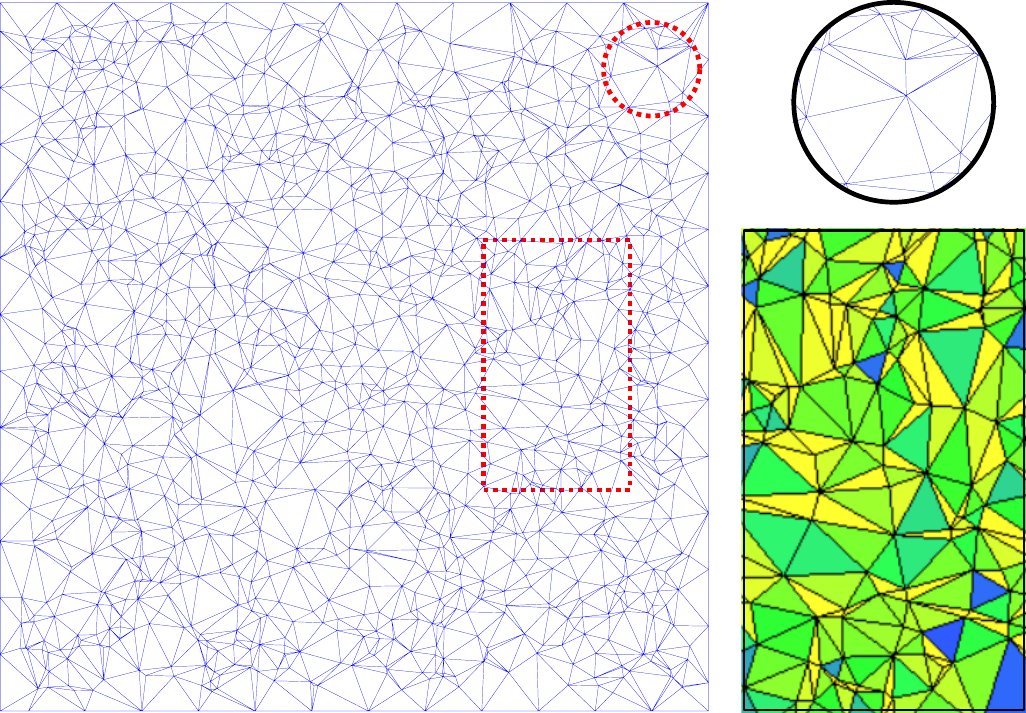}\label{square_b}}\qquad
    \subfloat[Circle mesh before smoothing]{\includegraphics[width=0.45\columnwidth]{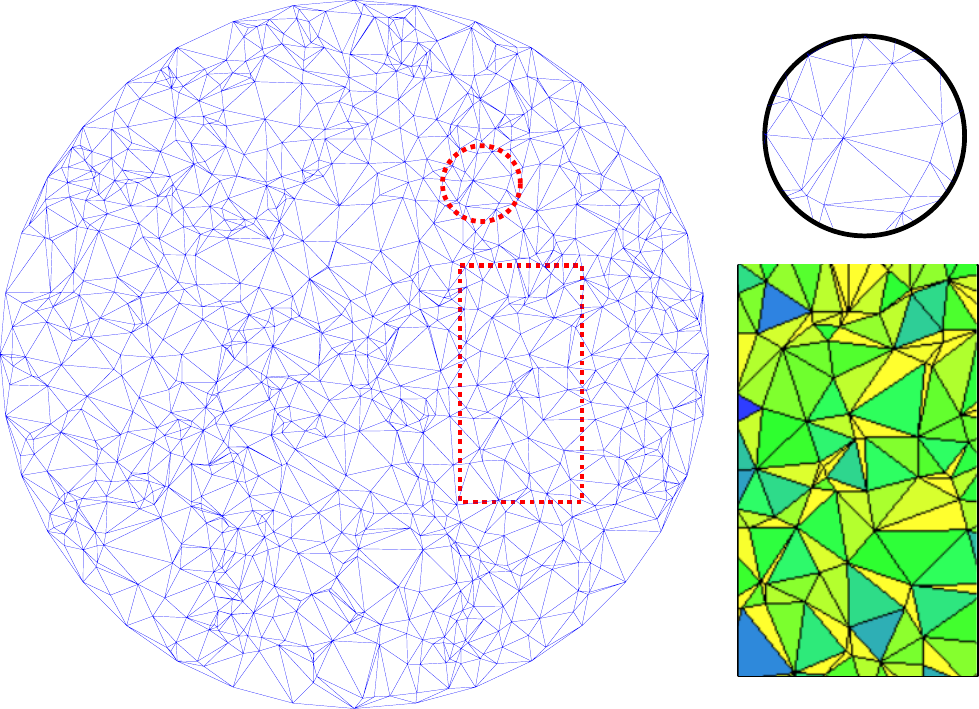}\label{circle_b}}\qquad 
    \subfloat[Airfoil mesh before smoothing]{\includegraphics[width=0.45\columnwidth]{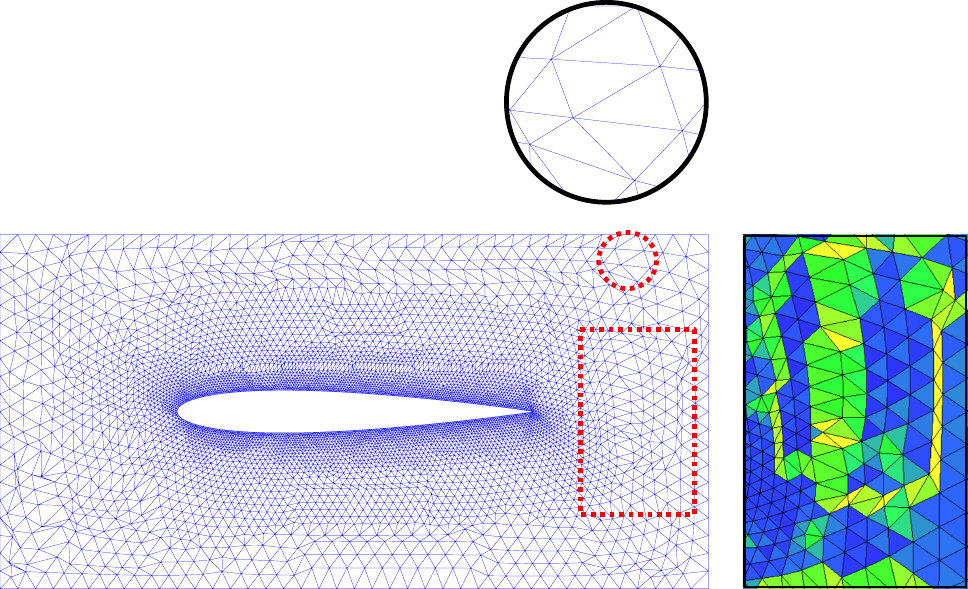}\label{airfoil_b}}\qquad 
    \subfloat[Pipe mesh before smoothing]{\includegraphics[width=0.45\columnwidth]{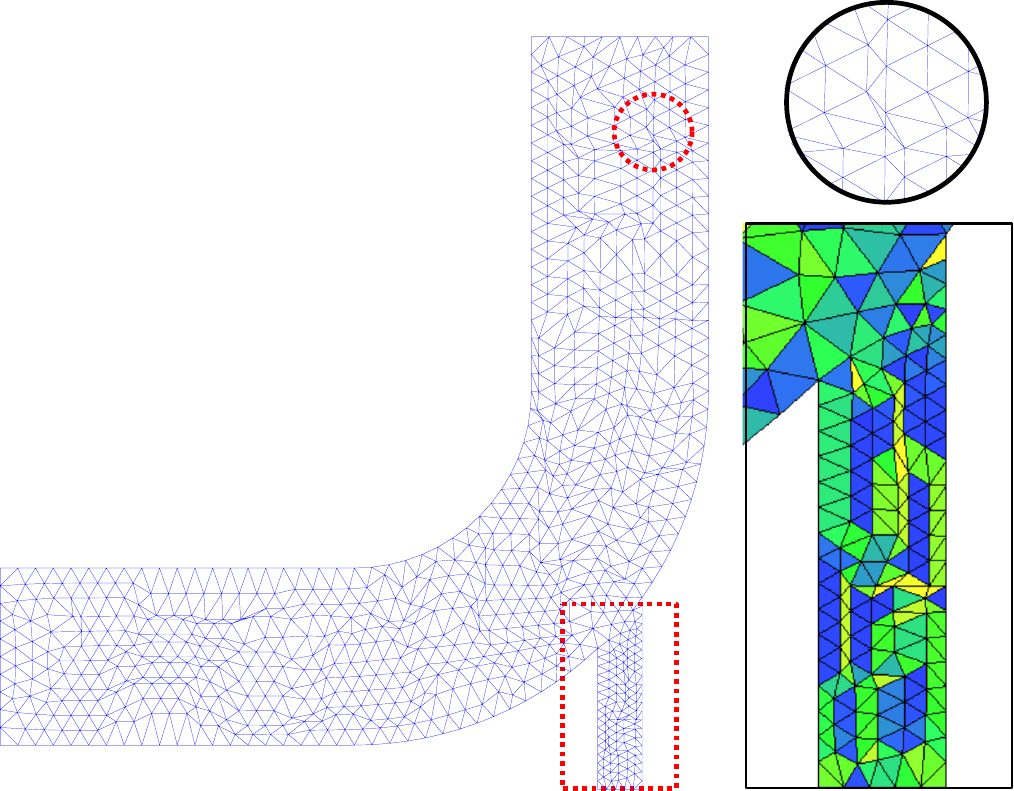}\label{pipe_b}}\qquad \\
    \subfloat[Square mesh after smoothing]{\includegraphics[width=0.45\columnwidth]{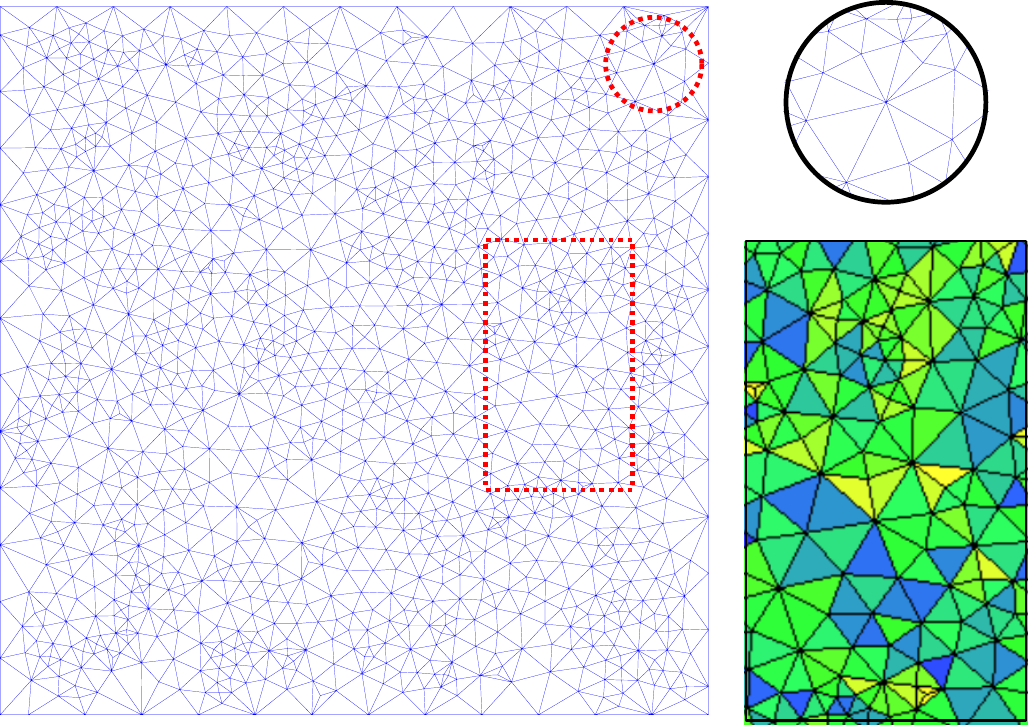}\label{square_a}}\qquad
    \subfloat[Circle mesh after smoothing]{\includegraphics[width=0.45\columnwidth]{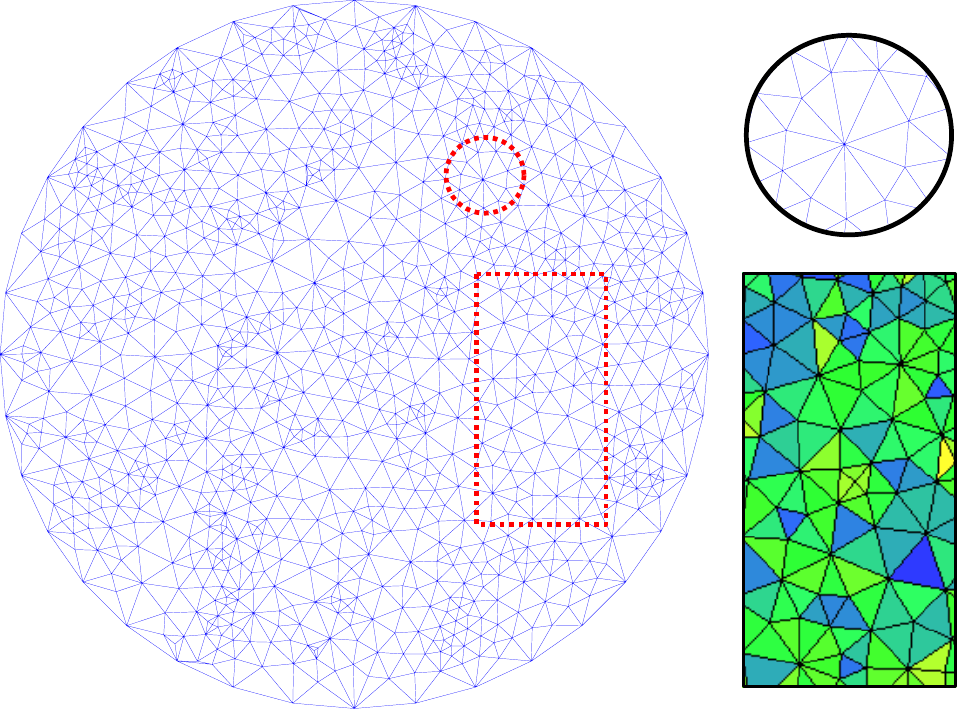}\label{circle_a}} \qquad
    \subfloat[Airfoil mesh after smoothing]{\includegraphics[width=0.45\columnwidth]{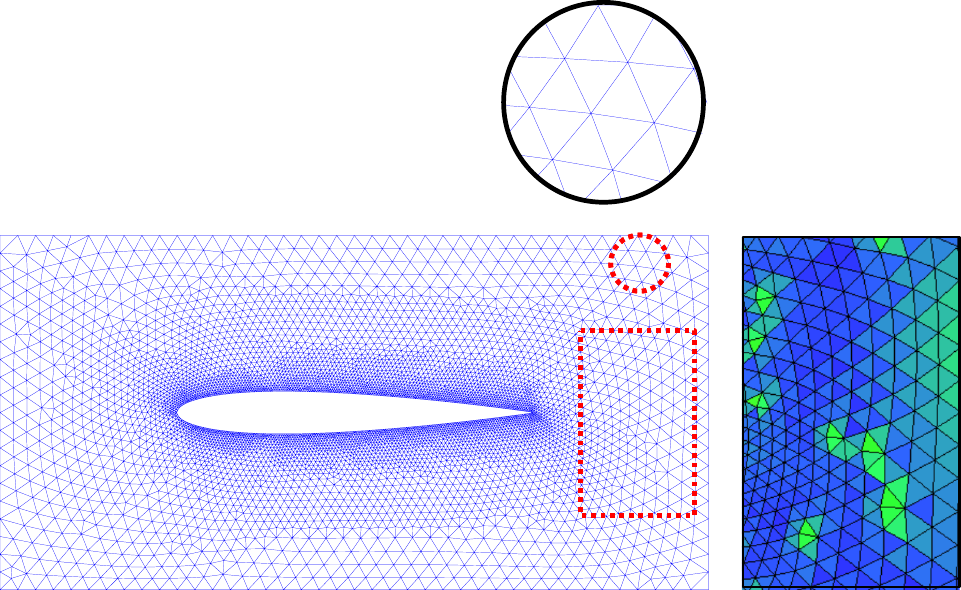}\label{airfoil_a}} \qquad
    \subfloat[Pipe mesh after smoothing]{\includegraphics[width=0.45\columnwidth]{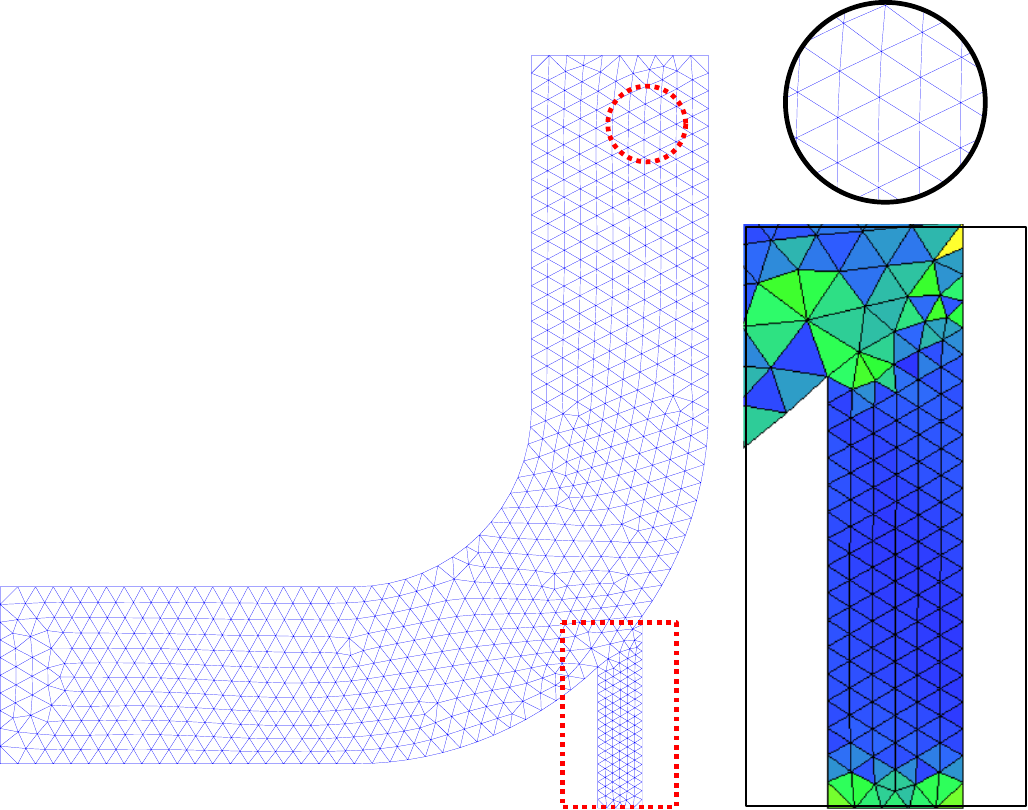}\label{pipe_a}}
    \caption{Mesh smoothing results of GMSNet on the test cases. Mesh nodes in meshes a) and b) are randomly generated in the domain, and mesh nodes in meshes c) and d) are manually adjusted to introduce distorted elements. \replace{High-quality elements are colored blue, and low-quality elements are colored yellow.}{In the detailed view, we utilize a color scheme to denote the quality of the mesh. Mesh elements range from blue for higher quality to yellow for lower quality, as indicated in the legend.}}\label{exp_cases}
\end{figure*}

We conducted tests on four mesh cases to evaluate the model's smoothing performance. To test the performance of the model more thoroughly, we constructed meshes containing highly distorted elements as test cases, as shown in the first row of Figure \ref{exp_cases}. For the first two meshes, mesh nodes were generated by uniform sampling within the geometric domain. The latter two meshes were initially created using meshing software to generate high-quality meshes, and then manual adjustments were made to introduce distorted mesh elements. It is worth noting that the meshes we tested are not included in our dataset, which solely consists of randomly generated two-dimensional unstructured  meshes with a square domain.

To facilitate a fair comparison among different models, we have implemented the algorithms within the same framework. The variations among the algorithms lie in how they generate the optimized point positions. It is essential to mention that we have employed sequential  algorithms in our study. However, for simpler algorithms like Laplacian smoothing, updating all nodes simultaneously is straightforward and fast. We have chosen the minimum angle, maximum angle, and the reciprocal of the aspect ratio as evaluation metrics for mesh quality. We conducted ten experimental runs for each case, with a maximum of 100 smoothing iterations per experiment. The best-smoothed mesh was selected as the final result based on the  weighted quality metric, which is defined as:
\begin{equation}
    \begin{aligned}
        \hat{q}&=\frac{1}{6}[\frac{\alpha_{\text{mean}}+\alpha_{\text{min}}+120-\beta_{\text{max}}-\beta_{\text{mean}}}{60}\\
        &+\left(\frac{1}{q}\right)_{\text{mean}}+\left(\frac{1}{q}\right)_{\text{min}}].
    \end{aligned}
\end{equation}
where $\alpha$ is minimum angle, $\beta$ is the maximum angle, and $q$ is the aspect ratio. When evaluating mesh elements, a smaller maximum angle and aspect ratio are preferable, while a larger minimum angle is desirable. For an ideal mesh element, the maximum angle, minimum angle, and aspect ratio should be 60 degrees, 60 degrees, and 1, respectively. The algorithm speed is measured in terms of the time taken to process a single mesh node. The results of mesh smoothing using the GMSNet are depicted in the second column of Figure  \ref{exp_cases}, and a comprehensive comparison of the performances of each algorithm is summarized in Table \ref{tab:mesh_quality}. 
\begin{table*}[htb]\footnotesize
    \centering
    \caption{Performance  of mesh smoothing algorithms. The best results are highlighted in bold, while the second-best results are annotated with a gray background.}
    \label{tab:mesh_quality}
    \begin{tabular}{ccccccccc}
        \toprule
        & & \multicolumn{2}{c}{Min. Angle }& \multicolumn{2}{c}{Max. Angle } & \multicolumn{2}{c}{ $\frac{1}{\text{Aspect ratio}}$} &  s/per node  \\
        \cmidrule{3-4} \cmidrule{5-6} \cmidrule{7-8}
        Mesh & Algorithm & min & mean & max & mean & min & mean &  \\
        \midrule
        \multirow{7}{*}{Square} & Origin & 0.04 & 29.65 & 179.82 & 95.09 & 0.00 & 0.58 &-   \\
        & LaplacianSmoothing & 12.38 & \textbf{43.97} & {147.67} & \textbf{79.44} & \colorbox{gray!40}{0.25} & \textbf{0.79} & 1.28E-04
        \\
        & AngleSmoothing & 10.52 & 41.26 & 148.6 & 81.76 & 0.22 & \colorbox{gray!40}{0.76} & 5.66E-04
        \\
        
        & CVTSmoothing & 2.55 & 40.88 & 170.6 & 84.04 & 0.05 & 0.75 & 1.17E-03
        \\
        & GETMeSmoothing &13.30 &	36.30 &	\colorbox{gray!40}{146.88 }	&87.63 &	\colorbox{gray!40}{0.25} &	0.69    & 2.10E-03
        \\

        & OptimSmoothing & \textbf{16.33} & \colorbox{gray!40}{43.94} & \textbf{136.99} & 80.7 & \textbf{0.32} & \textbf{0.79} & 9.27E-03
        \\
        
        & NN-Smoothing & 9.14 & 43.58 & 158.97 & \colorbox{gray!40}{79.79} & 0.16 & \textbf{0.79} & 4.67E-04
        \\
        
        & GMSNet & \colorbox{gray!40}{13.99} & 43.81 & 151.39 & 79.81 & 0.22 & \textbf{0.79} & 6.39E-04
        \\
        \midrule
        \multirow{7}{*}{Circle} &Origin & 0.29 & 30.38 & 179.05 & 94.60 & 0.01 & 0.59 & -  \\
        & LaplacianSmoothing & 1.43 & 42.86 & 174.53 & \textbf{81.19} & 0.03 & \textbf{0.78} & 1.31E-04
        \\
        
        & AngleSmoothing & 1.95 & 39.47 & 173.97 & 84.23 & 0.04 & 0.73 & 5.95E-04
        \\
        
        & CVTSmoothing & \colorbox{gray!40}{3.19} & 39.98 & 172.98 & 85.59 & \colorbox{gray!40}{0.05} & 0.73 & 1.20E-03
        \\
        & GETMeSmoothing &1.17 &	36.83 &	172.90 	&88.08 &	0.03 &	0.69 & 2.18E-03
        \\

        & OptimSmoothing & \textbf{6.52} & \textbf{43.31} & \textbf{154.66} & 81.9 & \textbf{0.15} & \textbf{0.78} & 8.78E-03
        \\

        & NN-Smoothing & 1.26 & 41.1 & 176.04 & 83.86 & 0.03 & \colorbox{gray!40}{0.75} & 4.83E-04
        \\
        
        & GMSNet & 2.2 & \colorbox{gray!40}{43} & \colorbox{gray!40}{169.21} & \colorbox{gray!40}{81.29} & \colorbox{gray!40}{0.05} & \textbf{0.78} & 6.58E-04
        \\
        \midrule
        \multirow{7}{*}{Airfoil} & Origin & 0.25 & 53.25 & 178.91 & 67.84 & 0.01 & 0.92 & -   \\
        & LaplacianSmoothing & 26.36 & \textbf{54.91} & 111.02 & \colorbox{gray!40}{65.78} & 0.51 & \textbf{0.94} & 1.34E-04         \\
        
        & AngleSmoothing & 20.26 & 53.49 & 118.76 & 66.64 & 0.42 & \colorbox{gray!40}{0.93} & 6.11E-04 \\
        
        & CVTSmoothing & 25.49 & 52.08 & 115.79 & 68.19 & 0.48 & 0.91 & 1.22E-03         \\
        & GETMeSmoothing &\textbf{36.02} &	52.90 	& \textbf{95.78 }	&67.89 &	\textbf{0.65 }	&0.92 & 2.24E-03         \\
        
        & OptimSmoothing & \colorbox{gray!40}{30.53} & \colorbox{gray!40}{54.9} & \colorbox{gray!40}{108.85} & \textbf{65.72} & \colorbox{gray!40}{0.54} & \textbf{0.94} & 8.40E-03\\
        
        & NN-Smoothing & {27.69} & 54.06 & 113.06 & 66.57 & 0.51 & \colorbox{gray!40}{0.93} & 4.88E-04         \\
        
        & GMSNet & 27.17 & 54.5 & {110.47} & 66.08 & {0.52} & \textbf{0.94} & 6.71E-04
        \\
        \midrule
        \multirow{7}{*}{Pipe} & Origin & 4.10 & 46.28 & 170.48 & 76.81 & 0.07 & 0.82 & -  \\
        & LaplacianSmoothing & 27.91 & \textbf{56.55} & 112.45 & \textbf{63.93} & 0.52 & \textbf{0.96} & 1.30E-04
        \\
        
        & AngleSmoothing & 24.28 & 54.62 & {101.93} & 65.69 & 0.53 & 0.94 & 5.81E-04
        \\
        
        & CVTSmoothing & 27.78 & 54.22 & \colorbox{gray!40}{97.8} & 66.28 & \colorbox{gray!40}{0.62} & 0.93 & 1.18E-03
        \\
        
        & GETMeSmoothing &\textbf{33.76 }	&51.07 &	\textbf{93.51} &	69.61 	&\textbf{0.65 }		&0.90   &    2.14E-03         
        \\
        
        & OptimSmoothing & \colorbox{gray!40}{32.39} & \colorbox{gray!40}{56.41} & 106.07 & \colorbox{gray!40}{64.14} & {0.57} & \textbf{0.96} & 9.01E-03
        \\
        
        & NN-Smoothing & {28.28} & 53.72 & 112.79 & 66.69 & 0.51 & 0.93 & 4.74E-04
        \\
        
        & GMSNet & 28.23 & 55.76 & 112.27 & 64.71 & 0.52 & \colorbox{gray!40}{0.95} & 6.47E-04
        \\
        \bottomrule
    \end{tabular}
\end{table*}

From Figure \ref{exp_cases}, it can be observed that for all four test cases, our proposed model significantly improves the mesh element quality. It is worth noting that GMSNet successsully smoothes mesh cases that are unseen during training (including circle mesh, airfoil mesh and pipe mesh). For meshes with reasonably distributed node degrees, our approach produces very smooth meshes, as shown in Figure \ref{airfoil_a} and \ref{pipe_a}. Moreover, our algorithm ensures robustness for highly distorted meshes, as demonstrated in Figure \ref{square_a} and \ref{circle_a}.
As shown in Table \ref{tab:mesh_quality}, our proposed algorithm outperforms most heuristic mesh smoothing methods, and the mesh element quality metrics closely approximate the results obtained using optimization-based algorithms. Since we adopted MetricLoss as the objective function for model optimization, our model has achieved the best or near-best performance in terms of the mean \textit{1/Aspect ratio} compared to other methods. When the mesh topology is reasonable, GETMeSmoothing exhibits a significant advantage in improving the minimum angle of the mesh. This is due to its use of a min-heap  to determine the order of smoothing for mesh elements, prioritizing the processing of the worst elements each time. However, in cases where the worst elements cannot be effectively addressed, its performance is not as strong as node-based methods, as observed in square and circle meshes. Although GMSNet does not make any assumptions about the order of smoothing nodes, it also performs well on highly distorted meshes. 

\begin{figure*}[htb]
    \centering
    \subfloat{\includegraphics[width=0.6\columnwidth]{Legend.pdf}}\\ \setcounter{subfigure}{0}
    \subfloat[Square mesh before smoothing]{\includegraphics[width=.6\columnwidth]{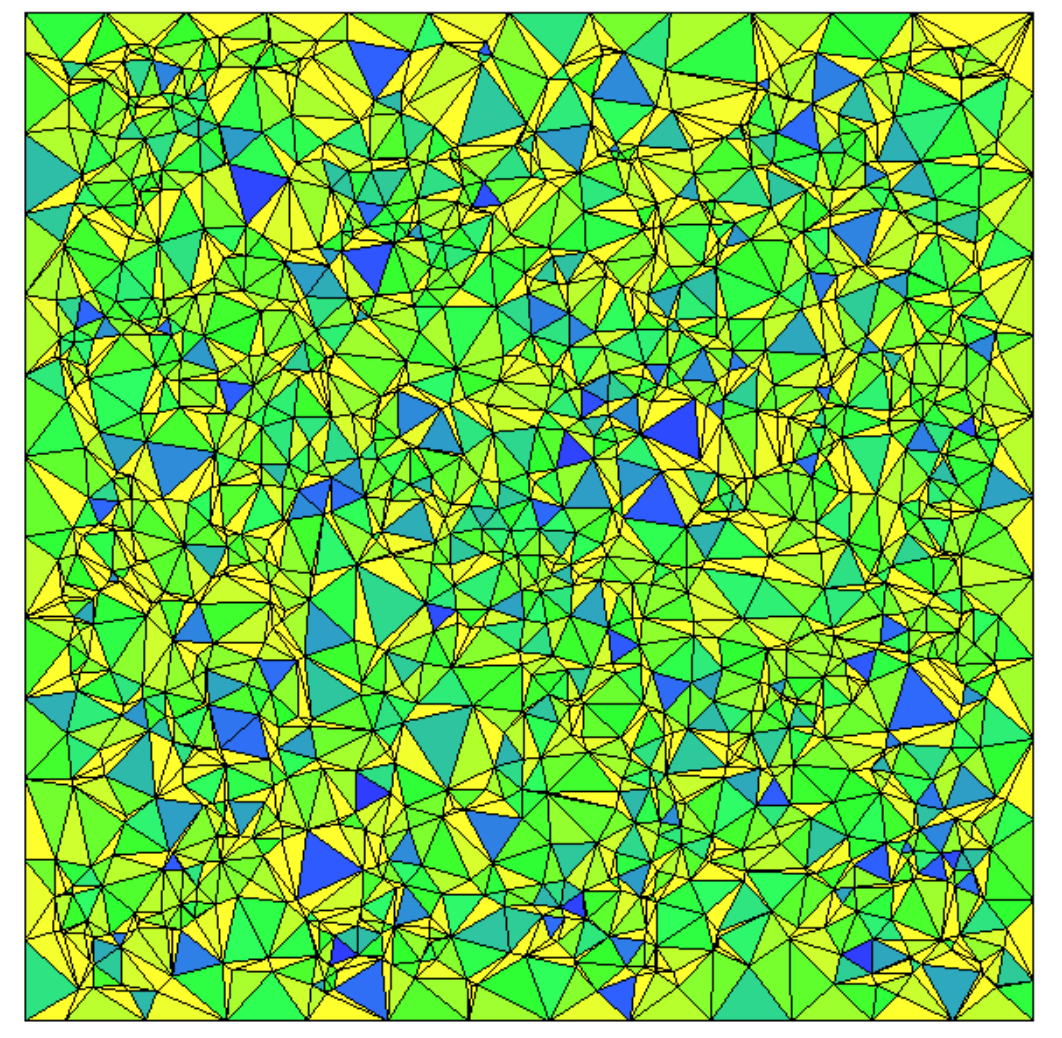}}\hspace{10pt}
    \subfloat[Square mesh smoothed by NN-Smoothing]{\includegraphics[width=.6\columnwidth]{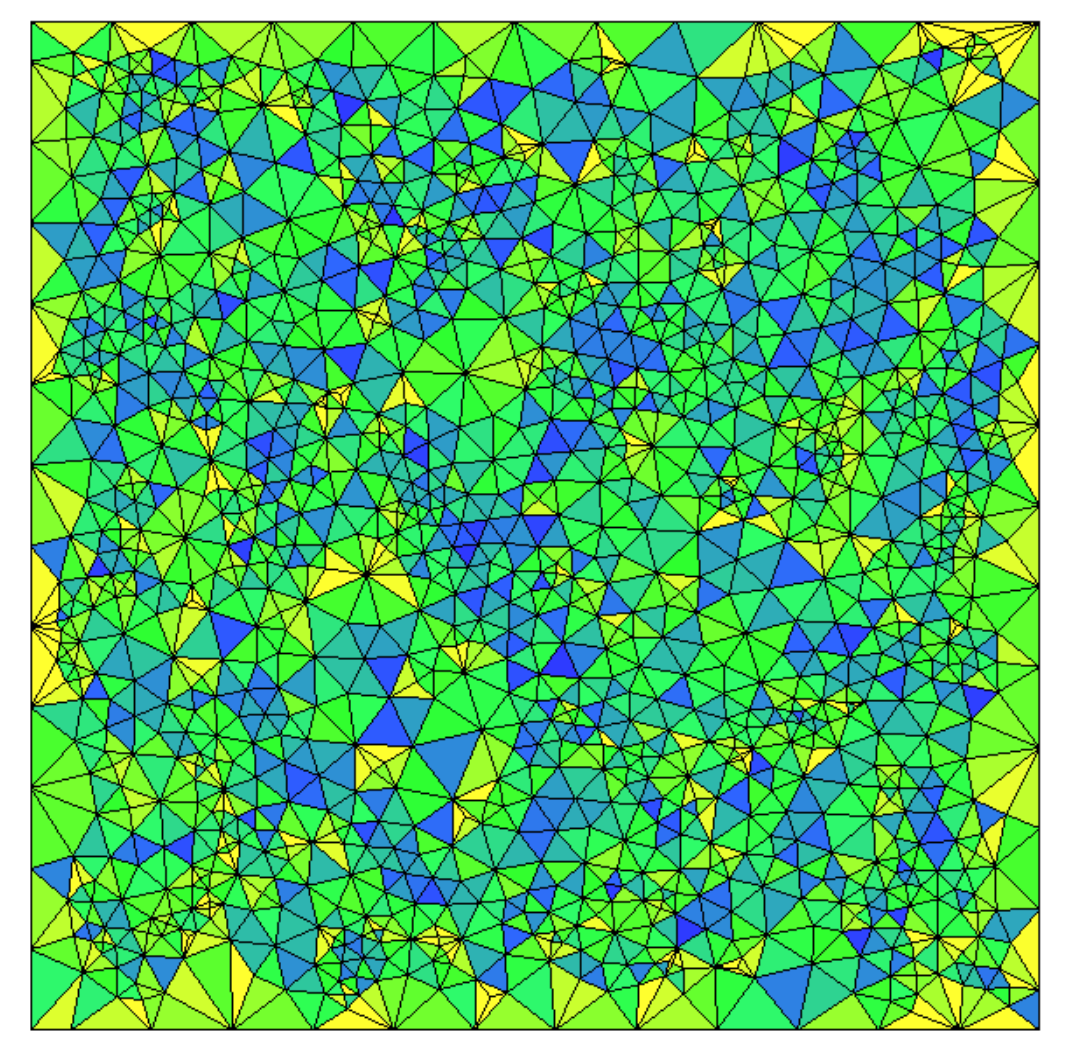}}\hspace{10pt}
    \subfloat[Square mesh smoothed by GMSNet]{\includegraphics[width=.6\columnwidth]{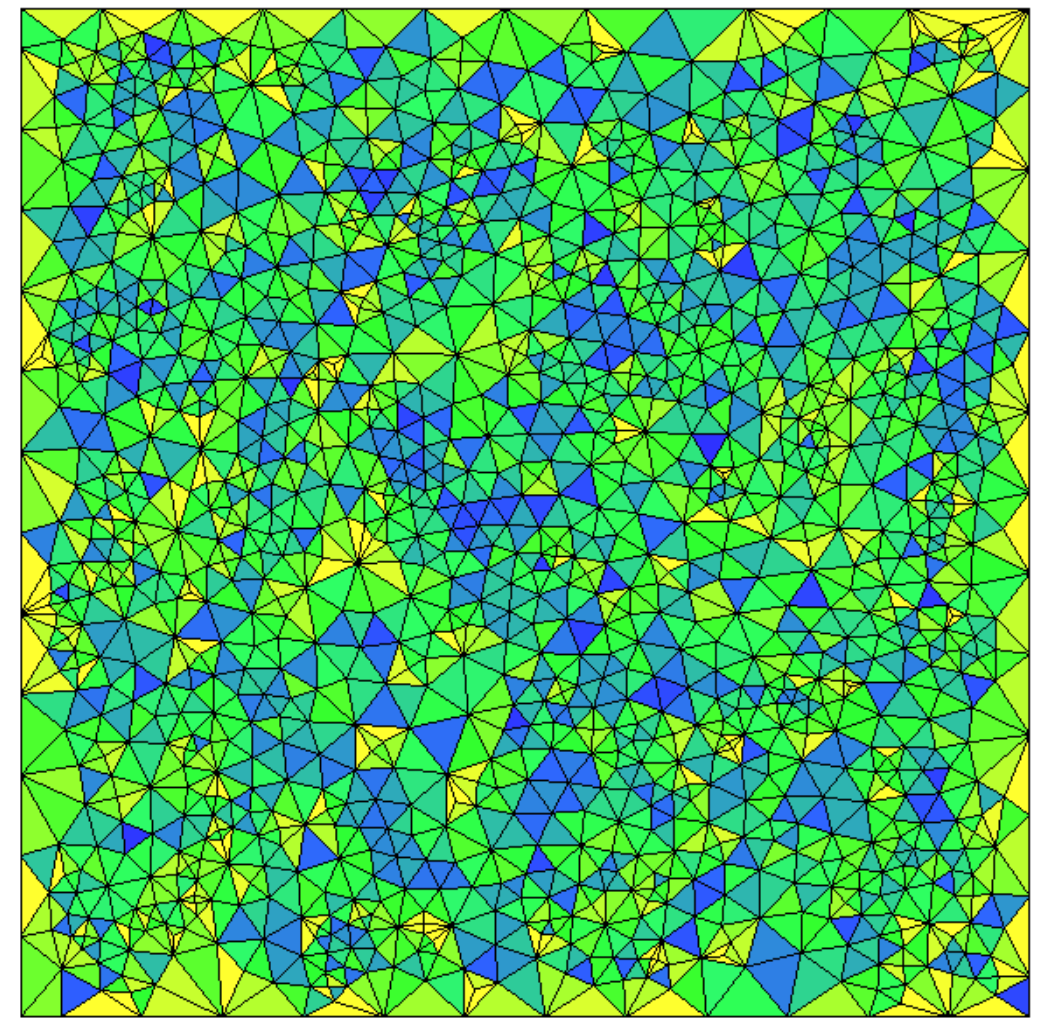}}\\
    \subfloat[Pipe mesh before smoothing]{\includegraphics[width=.6\columnwidth]{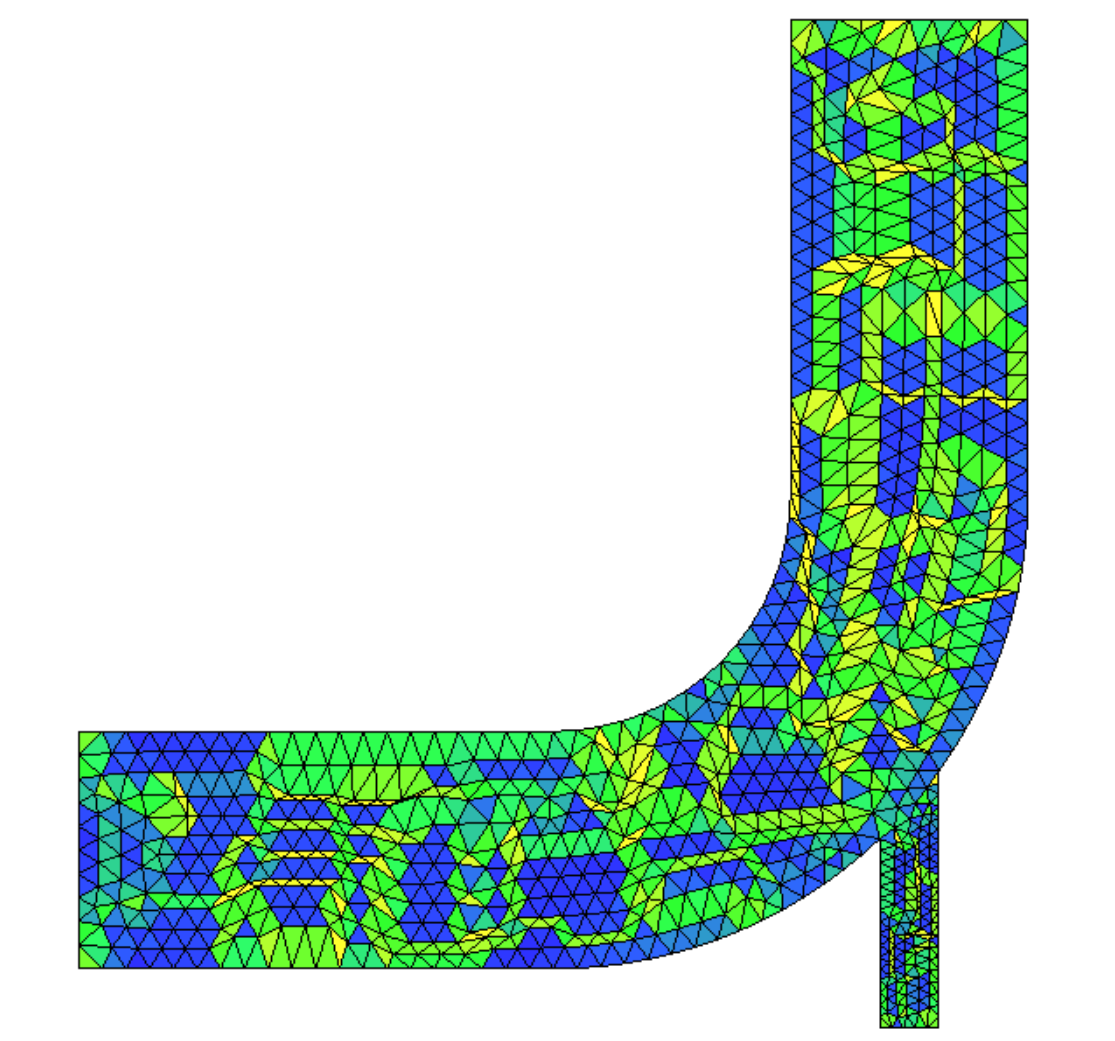}}\hspace{10pt}
    \subfloat[Pipe mesh smoothed by NN-Smoothing]{\includegraphics[width=.6\columnwidth]{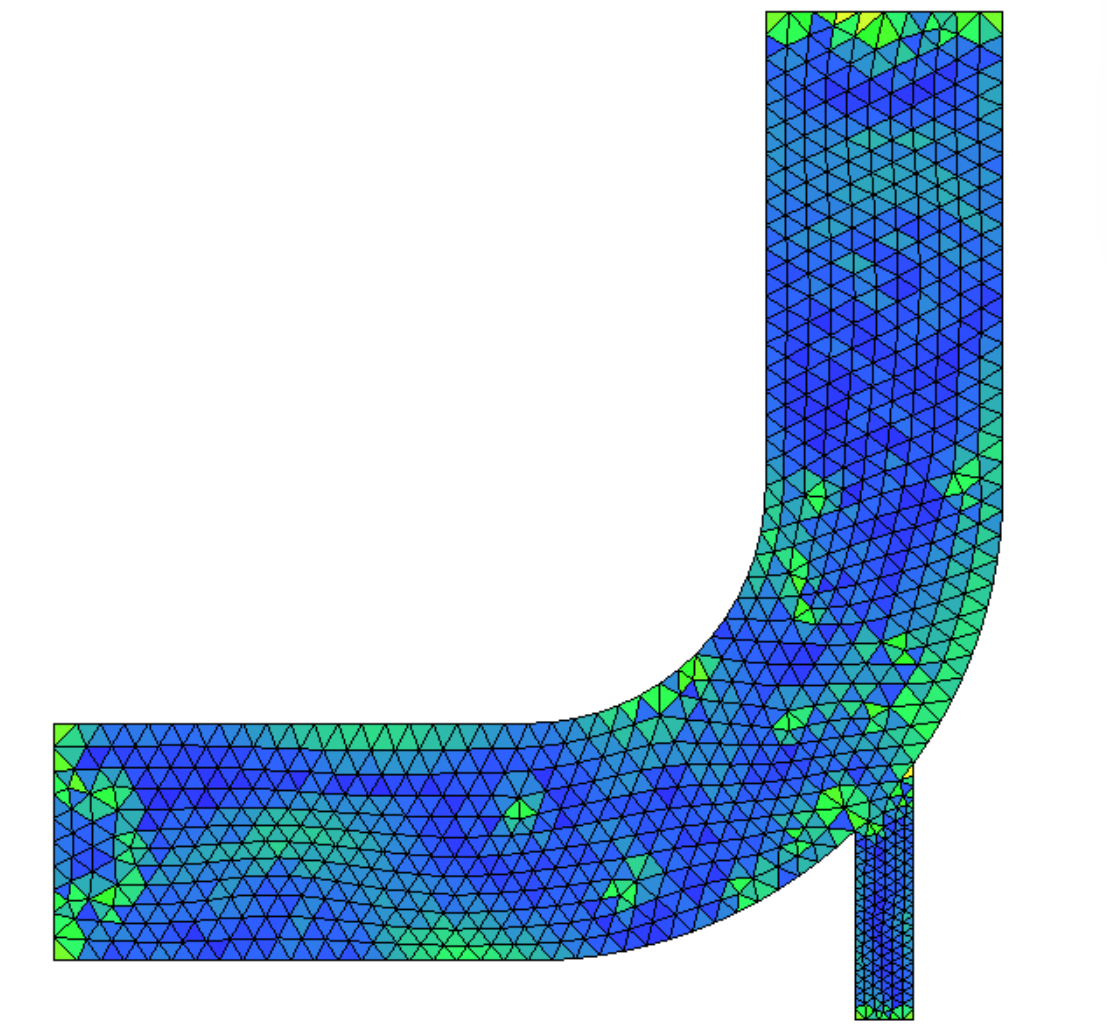}}\hspace{10pt}
    \subfloat[Pipe mesh smoothed by GMSNet]{\includegraphics[width=.6\columnwidth]{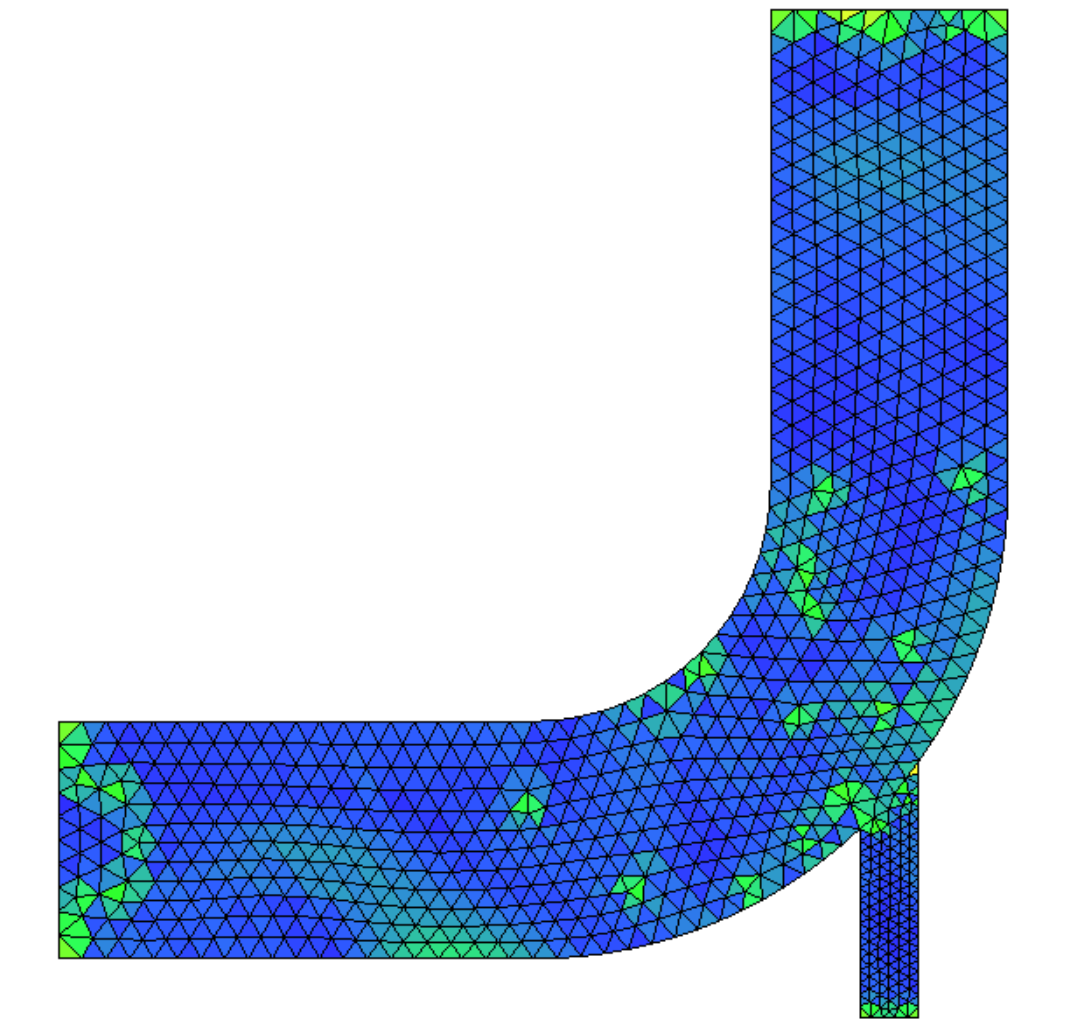}}
    \caption{Comparsion between GMSNet and NN-Smoothing. \add{The same color scheme as Figure 8 has been applied here.}}\label{GNNvsNN}
\end{figure*}

For all test cases, our model generally outperforms the NN-Smoothing model, despite the fact that we trained only one model to smooth nodes of different degrees, whereas NN-Smoothing requires training seven models.  Our model has only 5\% of the parameters compared to the NN-Smoothing model. The comparison of the smoothing performance of the two models is illustrated in Figure \ref{GNNvsNN}. It can be observed that the smoothing performance of both models are nearly identical. For the mesh with a poorer topology, the square mesh, both models significantly improve the orthogonality of the mesh elements. Additionally, GMSNet provides a slightly better distribution of mesh density. From the efficiency perspective, the proposed method exhibits a average speed improvement of  of 13.56 times compared to optimization-based approaches. Compared to the simple MLP  in NN-Smoothing, our model incorporates additional network modules, such as node encoders, graph convolutional layers, regularization layers, and more. These modules inevitably introduce additional computational overhead compared to NN-Smoothing. However, we believe that by adjusting hyperparameters \citep{tune}, optimizing network modules, and performing serialization operations \citep{torchjit}, our model should be able to achieve higher performance. We leave this for future work in the engineering of this method.

\begin{figure}[htb]
    \centering
    \subfloat[Sqaure mesh]{\includegraphics[width=.5\columnwidth]{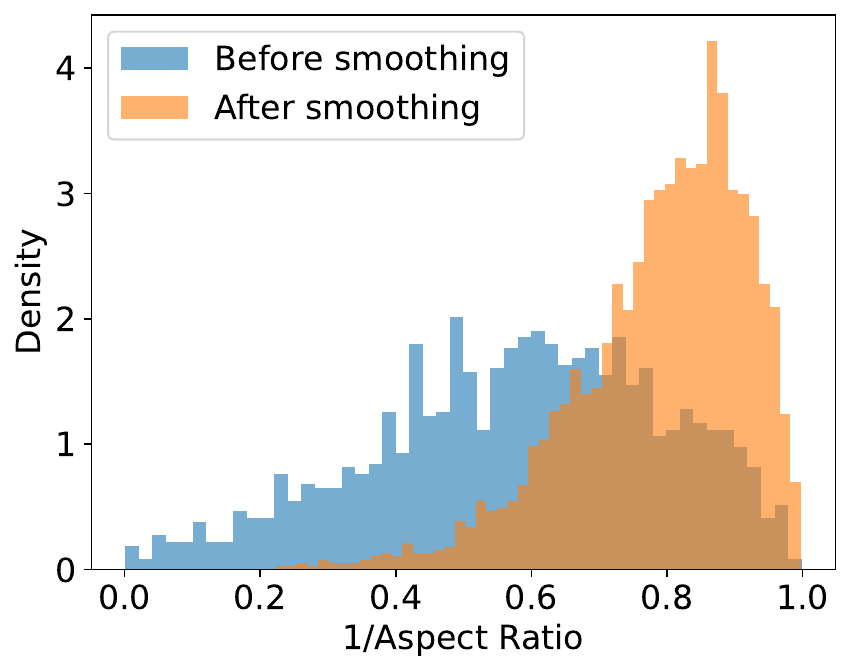}}
    \subfloat[Circle mesh]{\includegraphics[width=.5\columnwidth]{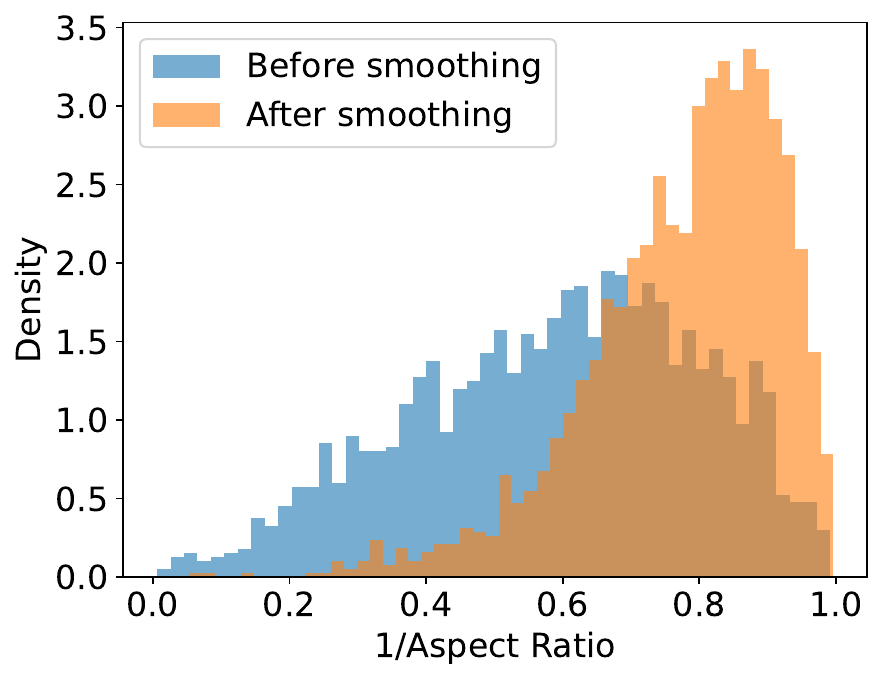}}\\
    \subfloat[Airfoil mesh]{\includegraphics[width=.5\columnwidth]{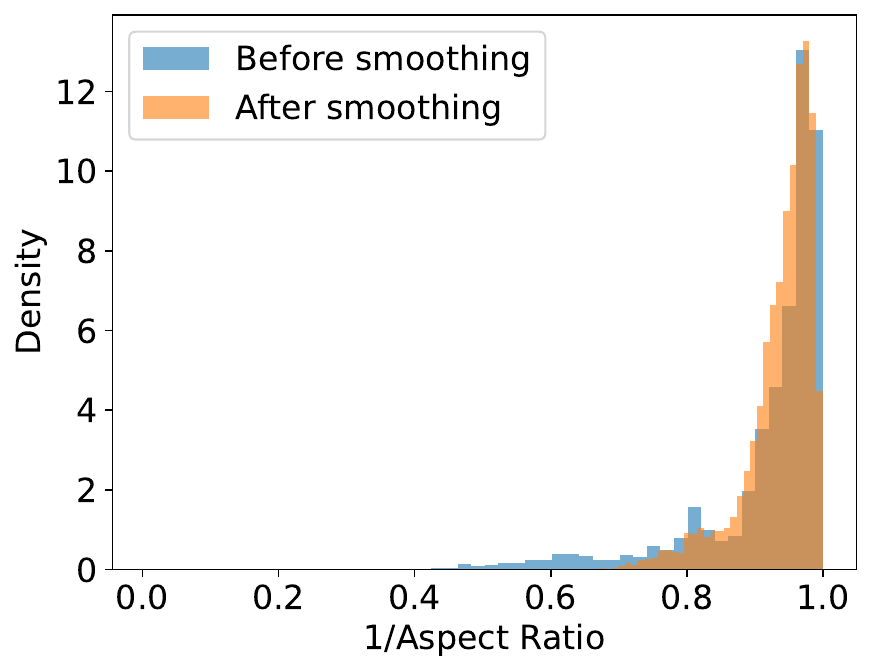}}
    \subfloat[Pipe mesh]{\includegraphics[width=.5\columnwidth]{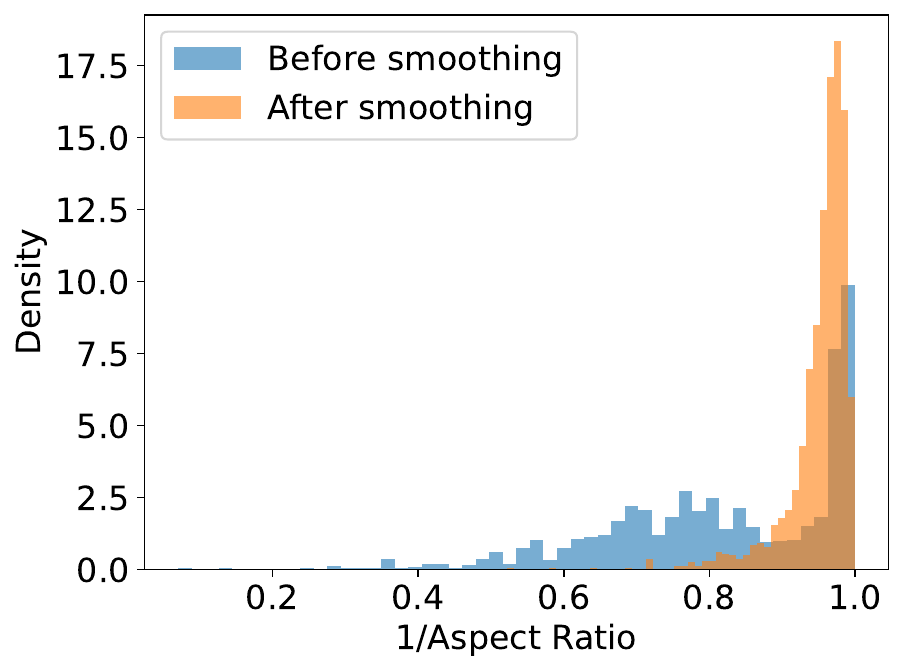}}\\
    \caption{The quality distribution of mesh elements before and after GMSNet smoothing.}\label{fig:metricdist}
\end{figure}

We also validate the effectiveness of our proposed algorithm by examining the distribution of mesh element quality. The  quality distributions of mesh elements before and after smoothing are shown in Figure \ref{fig:metricdist}. It can be observed that the algorithm significantly increases the proportion of high-quality mesh elements and improves the overall mesh quality. In summary, the experimental results demonstrate the effectiveness and efficiency of our proposed model for mesh smoothing tasks.
\subsection{The influence of different loss functions}\label{sec:loss_func}
In Section \ref{sec:loss_design}, we designed a loss function for model training. In this section, we discuss the effectiveness of MetricLoss and how to choose the appropriate loss function. We compare the performance of commonly used mesh quality metrics as loss  functions during model training, including:
\begin{itemize}
    \item Min-max angle loss:\\ 
    $ \mathcal{L}(\boldsymbol{x},\mathbf{S}(\boldsymbol{x}))=\frac{1}{|\mathbf{S}(\boldsymbol{x})|}\sum_{i=1}^{|\mathbf{S}(\boldsymbol{x})|}{\min\max(\theta_{ij})}$, {for $j=1,2,3$}, where $\theta_{ij}$ is the angle in triangle $T_i$.
    \item Aspect ratio loss:\\ $\mathcal{L}(\boldsymbol{x},\mathbf{S}(\boldsymbol{x}))=\frac{1}{|\mathbf{S}(\boldsymbol{x})|}\sum_{i=1}^{|\mathbf{S}(\boldsymbol{x})|}{\frac{m_i^2+
            n_i^2+l_i^2}{4\sqrt{3} S_i}} $,  where the symbols are defined in Section \ref{sec:loss_design}.
    \item Cosine loss:\\ $\mathcal{L}(\boldsymbol{x},\mathbf{S}(\boldsymbol{x}))=\frac{1}{3|\mathbf{S}(\boldsymbol{x})|}\sum_{i=1}^{|\mathbf{S}(\boldsymbol{x})|}{(\cos(\theta_{ij})-\frac{1}{2})^2} $
    \item MetricLoss: \\
    $	\mathcal{L}(\boldsymbol{x},\mathbf{S}(\boldsymbol{x}))=\frac{1}{|\mathbf{S}(\boldsymbol{x})|}\sum_{i=1}^{|\mathbf{S}(\boldsymbol{x})|}{(1-\frac{4\sqrt{3} S_i}{m_i^2+
            n_i^2+l_i^2} )}$
\end{itemize}

\begin{figure}[htb]
    \centering
    \subfloat[Aspect ratio loss]{\includegraphics[width=.48\columnwidth]{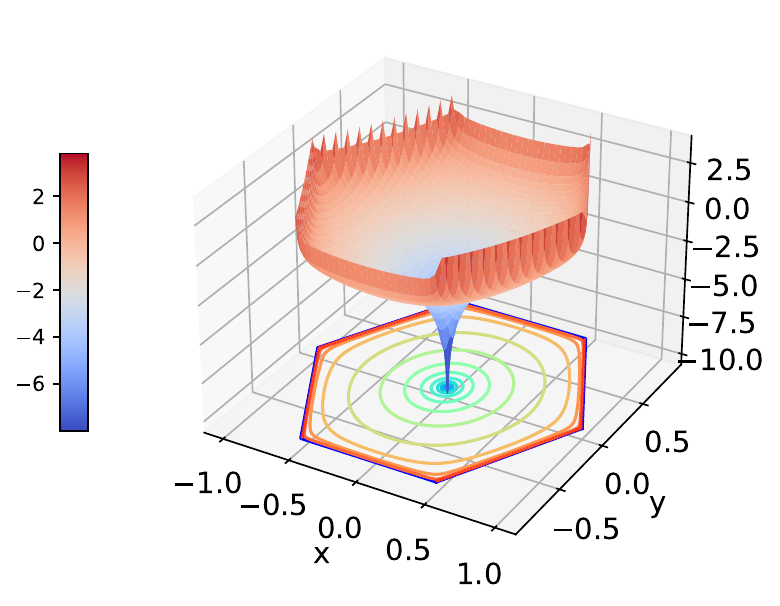}\label{fig:distort_loss}}\hspace{2pt}
    \subfloat[Min-max angle loss]{\includegraphics[width=.48\columnwidth]{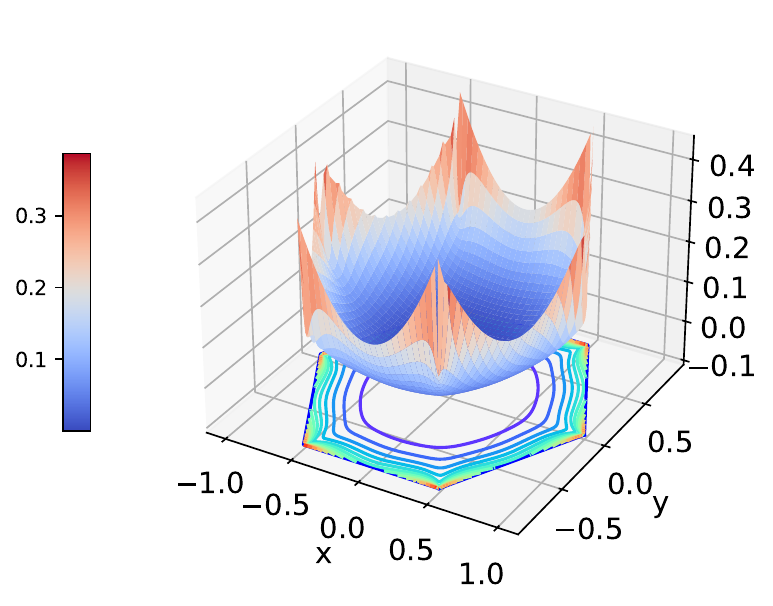}}\\
    \subfloat[Cosine loss]{\includegraphics[width=.48\columnwidth]{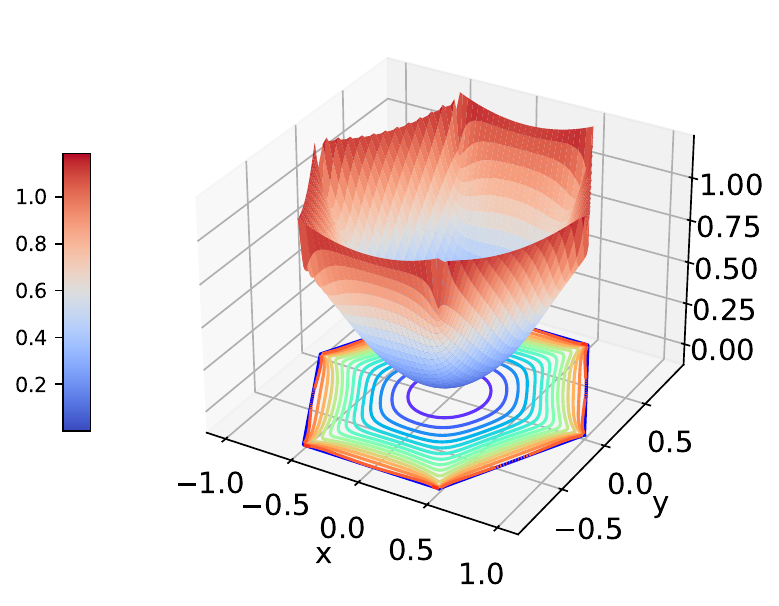}\label{fig:cos_loss}}\hspace{2pt}
    \subfloat[The proposed loss]{\includegraphics[width=.48\columnwidth]{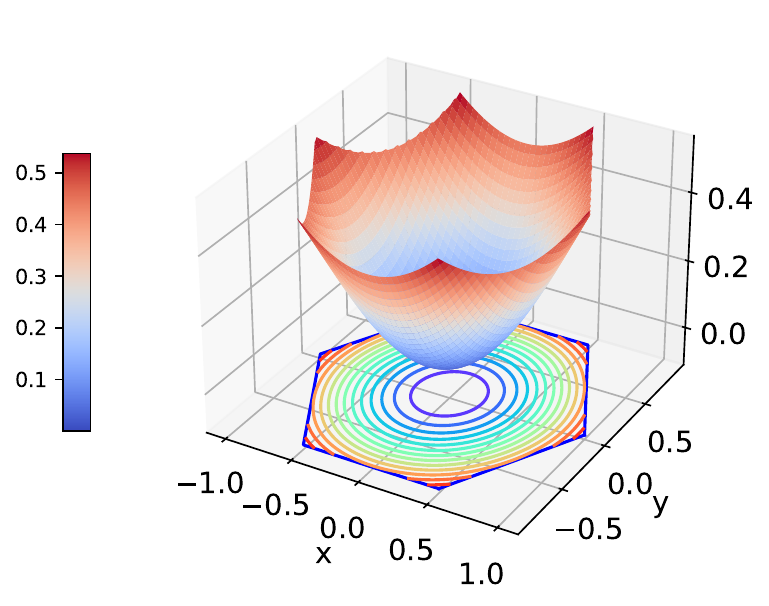}}\\
    \caption{Plots of different loss functions. The range and smoothness of the loss function influence the model's convergence. In Figure \ref{fig:distort_loss}, The z-coordinate is given as $log(z)$.}\label{fig:loss}
\end{figure}
\begin{figure}[htb]
    \centering
    \includegraphics[width=\linewidth]{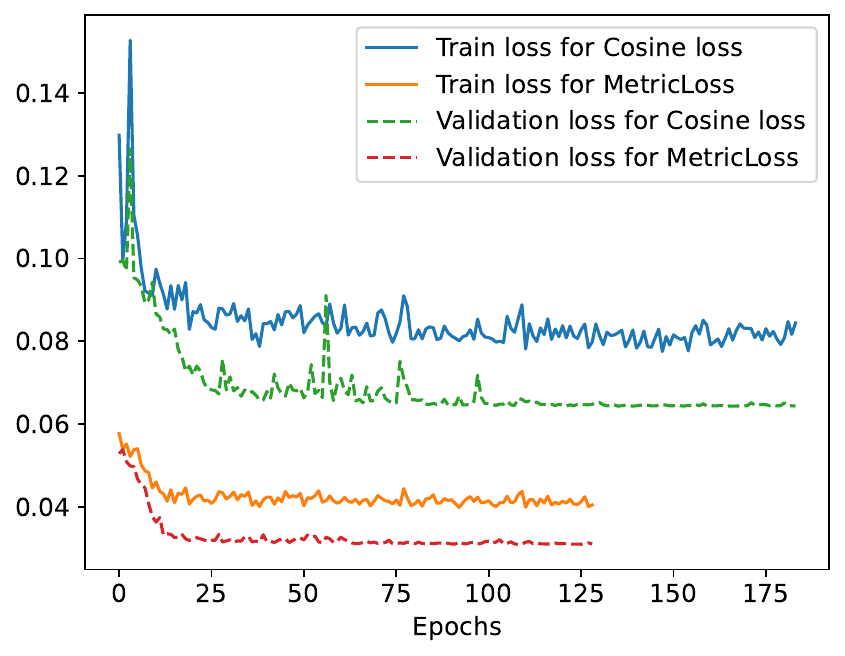}
    \caption{Comparison of  different loss functions. The loss functions based on min-max angle and aspect ratio encountered training failures during the initial epochs. In contrast, both MetricLoss and the transformed Cosine loss function yielded convergent results.}
    \label{fig:valloss}
\end{figure}

The  plots of the aforementioned loss functions with respect to the variations in the central node of the \textit{StarPolygon} are shown in Figure \ref{fig:loss}. It can be observed that the loss function we employed exhibits smoother transformations as the input changes, which facilitates the optimization process. We can also see than the aspect ratio loss and min-max angle loss are not suitable to be employed as the loss function. When encountering highly distorted elements, the former leads to numerical values approaching positive infinity, causing a sharp increase in loss and inducing training instability. The latter, on the other hand, results in abrupt changes in loss when facing highly distorted elements, while the loss function remains relatively flat when the central node is situated in the neighborhood of the optimal position. This circumstance also hinders the optimization process.  Although the original angle-based loss function is difficult to train, the loss function transformed using the cosine function is more stable, as shown in Figure~\ref{fig:cos_loss}. We trained the model using different loss functions, employing the same model configuration and training configuration. The train loss and validation loss of models with different loss functions are depicted in Figure \ref{fig:valloss}. It can be observed that MetricLoss and cosine loss function are suitable for training the model, as the model converges rapidly after several iterations. Moreover, MetricLoss is remarkably stable during the training process. The  experiments between two models based on different loss functions are shown in Table \ref{tab:loss_comparison}. It can be seen that the models show similar performance. 
\begin{table*}[htb]\footnotesize
    \centering
    \caption{Comparison of cosine loss and proposed loss on Pipe and Square meshes}
    \label{tab:loss_comparison}
    \begin{tabular}{cccccccccccc}
        \toprule
        & & \multicolumn{2}{c}{Min. Angle} & \multicolumn{2}{c}{Max. Angle} & \multicolumn{2}{c}{ $\frac{1}{\text{Aspect ratio}}$}   \\
        \cmidrule{3-4} \cmidrule{5-6} \cmidrule{7-8}
        Mesh & Algorithm & min & mean & max & mean & min & mean &  \\
        \midrule
        Pipe & Cosine loss & 26.77 & 55.71 & 110.43 & 64.67 & 0.54 & 0.95 \\
        & MetricLoss & 26.37 & 55.56 & 107.73 & 64.82 & 0.56 & 0.95 \\
        \midrule
        Square & Cosine loss & 11.04 & 43.76 & 146.87 & 79.61 & 0.25 & 0.79 \\
        & MetricLoss & 13.53 & 44.06 & 150.46 & 79.65 & 0.22 & 0.79 \\
        \bottomrule
    \end{tabular}
\end{table*}

When training with the other two loss functions, the model either fails to converge or exhibits severe oscillations. This indicates that employing  mesh quality  metrics with a larger range of numerical values or rapid changes in the numerical values when encountering highly distorted elements is detrimental to the training of the model. In summary, various loss functions can be used for  mesh smoothing model training. When constructing the loss function through mesh quality metrics, it is necessary to consider the behavior of the function when encountering distorted elements, and to avoid selecting loss functions with a larger range of numerical values or abrupt changes. Inappropriate mesh quality metrics can also be transformed and employed for model training, as demonstrated by  the min-max angle loss and cosine loss.

\section{Conclusion}\label{sec13}

In this paper, we propose a graph neural network model, GMSNet, for intelligent mesh smoothing. The proposed model takes the neighbours of mesh nodes as input and learns  to directly output the smoothed positions of mesh nodes, avoiding the computational overhead associated with the optim\-ization-based  smoothing. A fault-tolerance mechanism, shift truncation, is also applied to prevent negative volume elements. With a lightweight design, GMSNet can be applied to   mesh nodes with varying degrees, and is not affected by the data input order. We also introduce an novel  loss function, MetricLoss, based on mesh quality metrics, eliminating the need for high-quality mesh generation cost to train the model.   Experiments on  two-dimensional triangle meshes demonstrates that our proposed model achieves outstanding smoothing performance   with   an  average acceleration of 13.56 times compared to to optimization-based  smoothing. The results also show that GMSNet achieve superior performance than  NN-Smoothing model with just 5\% model parameters.  We also illustrate that  MetricLoss achieves fast and stable model training through experiments.

Future work includes applying our proposed method to other types of mesh elements and extending it to surface and volume meshes. Introducing edge flipping and mesh density modification into the mesh smoothing process to achieve superior mesh smoothing effects is also an approach worth exploring. Improving the model's efficiency to enable its application on large-scale meshes needs careful  consideration.

\bibliographystyle{fitee}
\bibliography{ref}

\begin{thebibliography}{}\footnotesize
\itemsep -4pt
\vspace{-8pt}

\bibitem[\protect\astroncite{Baker}{2005}]{c6}
Baker TJ, 2005.
\newblock Mesh generation: Art or science?
\newblock {\em Progress in Aerospace Sciences},  41(1):29-63.

\bibitem[\protect\astroncite{Bohn and Feischl}{2021}]{bohn2021recurrent}
Bohn J, Feischl M, 2021.
\newblock Recurrent neural networks as optimal mesh refinement strategies.
\newblock {\em Computers \& Mathematics with Applications},  97:61-76.

\bibitem[\protect\astroncite{Bridgeman {et~al.}}{2010}]{c2}
Bridgeman J, Jefferson B, Parsons SA, 2010.
\newblock The development and application of cfd models for water treatment
  flocculators.
\newblock {\em Advances in Engineering Software},  41(1):99-109.

\bibitem[\protect\astroncite{Cai {et~al.}}{2021}]{c37}
Cai T, Luo S, Xu K, {et~al.}, 2021.
\newblock Graphnorm: A principled approach to accelerating graph neural network
  training.
\newblock International Conference on Machine Learning, p.1204-1215.

\bibitem[\protect\astroncite{Canann
  {et~al.}}{1998}]{canannApproachCombinedLaplacian1998}
Canann SA, Tristano JR, Staten ML, 1998.
\newblock An {Approach} to {Combined} {Laplacian} and {Optimization}-{Based}
  {Smoothing} for {Triangular}, {Quadrilateral}, and {Quad}-{Dominant}
  {Meshes}.
\newblock {\em IMR},  1:479-94 (Publisher: Citeseer).

\bibitem[\protect\astroncite{Chen}{2004}]{chen2004mesh}
Chen L, 2004.
\newblock Mesh smoothing schemes based on optimal delaunay triangulations.
\newblock IMR, p.109-120.

\bibitem[\protect\astroncite{Chen {et~al.}}{2020}]{c27}
Chen X, Liu J, Pang Y, {et~al.}, 2020.
\newblock Developing a new mesh quality evaluation method based on
  convolutional neural network.
\newblock {\em Engineering Applications of Computational Fluid Mechanics},
  14(1):391-400.

\bibitem[\protect\astroncite{Chen {et~al.}}{2021}]{c29}
Chen X, Liu J, Gong C, {et~al.}, 2021.
\newblock Mve-net: An automatic 3-d structured mesh validity evaluation
  framework using deep neural networks.
\newblock {\em Computer-Aided Design},  141:103104.

\bibitem[\protect\astroncite{Chen {et~al.}}{2022}]{mgnet}
Chen X, Li T, Wan Q, {et~al.}, 2022.
\newblock Mgnet: a novel differential mesh generation method based on
  unsupervised neural networks.
\newblock {\em Engineering with Computers},  38(5):4409-4421.

\bibitem[\protect\astroncite{Chowdhary and Chowdhary}{2020}]{c26}
Chowdhary K, Chowdhary K, 2020.
\newblock Natural language processing.
\newblock {\em Fundamentals of artificial intelligence}, :603-649.

\bibitem[\protect\astroncite{Damjanovi{\'c} {et~al.}}{2011}]{c3}
Damjanovi{\'c} D, Kozak D, {\v{Z}}ivi{\'c} M, {et~al.}, 2011.
\newblock Cfd analysis of concept car in order to improve aerodynamics.
\newblock {\em J{\'a}rm{\H{u}}ipari innov{\'a}ci{\'o}},  1(2):108-115.

\bibitem[\protect\astroncite{Daroya {et~al.}}{2020}]{rein}
Daroya R, Atienza R, Cajote R, 2020.
\newblock Rein: Flexible mesh generation from point clouds.
\newblock Proceedings of the IEEE/CVF Conference on Computer Vision and Pattern
  Recognition Workshops, p.352-353.

\bibitem[\protect\astroncite{Darwish and Moukalled}{2016}]{c5}
Darwish M, Moukalled F, 2016.
\newblock The finite volume method in computational fluid dynamics: an advanced
  introduction with OpenFOAM{\textregistered} and Matlab{\textregistered}.
\newblock Springer, Berlin, p.110-128.

\bibitem[\protect\astroncite{Du and Gunzburger}{2002}]{c14}
Du Q, Gunzburger M, 2002.
\newblock Grid generation and optimization based on centroidal voronoi
  tessellations.
\newblock {\em Applied mathematics and computation},  133(2-3):591-607.

\bibitem[\protect\astroncite{Du and Wang}{2003}]{du2003tetrahedral}
Du Q, Wang D, 2003.
\newblock Tetrahedral mesh generation and optimization based on centroidal
  voronoi tessellations.
\newblock {\em International journal for numerical methods in engineering},
  56(9):1355-1373.

\bibitem[\protect\astroncite{Du {et~al.}}{1999}]{c16}
Du Q, Faber V, Gunzburger M, 1999.
\newblock Centroidal voronoi tessellations: Applications and algorithms.
\newblock {\em SIAM review},  41(4):637-676.

\bibitem[\protect\astroncite{Durand {et~al.}}{2019}]{durand2019general}
Durand R, Pantoja-Rosero B, Oliveira V, 2019.
\newblock A general mesh smoothing method for finite elements.
\newblock {\em Finite Elements in Analysis and Design},  158:17-30.

\bibitem[\protect\astroncite{Escobar {et~al.}}{2005}]{escobar2005smoothing}
Escobar J, Montenegro R, Montero G, {et~al.}, 2005.
\newblock Smoothing and local refinement techniques for improving tetrahedral
  mesh quality.
\newblock {\em Computers \& structures},  83(28-30):2423-2430.

\bibitem[\protect\astroncite{Fidkowski and Chen}{2021}]{fidkowski2021metric}
Fidkowski KJ, Chen G, 2021.
\newblock Metric-based, goal-oriented mesh adaptation using machine learning.
\newblock {\em Journal of Computational Physics},  426:109957.

\bibitem[\protect\astroncite{Field}{1988}]{field1988laplacian}
Field DA, 1988.
\newblock Laplacian smoothing and delaunay triangulations.
\newblock {\em Communications in applied numerical methods},  4(6):709-712.

\bibitem[\protect\astroncite{Freitag and Knupp}{2002}]{c9}
Freitag LA, Knupp PM, 2002.
\newblock Tetrahedral mesh improvement via optimization of the element
  condition number.
\newblock {\em International Journal for Numerical Methods in Engineering},
  53(6):1377-1391.

\bibitem[\protect\astroncite{Freitag and Ollivier-Gooch}{1997}]{c8}
Freitag LA, Ollivier-Gooch C, 1997.
\newblock Tetrahedral mesh improvement using swapping and smoothing.
\newblock {\em International Journal for Numerical Methods in Engineering},
  40(21):3979-4002.

\bibitem[\protect\astroncite{Guo and Hai}{2021}]{guo2021adaptive}
Guo Y, Hai Y, 2021.
\newblock Adaptive surface mesh remeshing based on a sphere packing method and
  a node insertion/deletion method.
\newblock {\em Applied Mathematical Modelling},  98:1-13.

\bibitem[\protect\astroncite{Guo {et~al.}}{2020}]{guo2020angle}
Guo Y, Wang L, Zhao K, {et~al.}, 2020.
\newblock An angle-based smoothing method for triangular and tetrahedral
  meshes.
\newblock Image and Graphics Technologies and Applications: 15th Chinese
  Conference, IGTA 2020, Beijing, China, September 19, 2020, Revised Selected
  Papers 15, p.224-234.

\bibitem[\protect\astroncite{Guo {et~al.}}{2021}]{c11}
Guo Y, Wang C, Ma Z, {et~al.}, 2021.
\newblock A new mesh smoothing method based on a neural network.
\newblock {\em Computational Mechanics}, :1-14.

\bibitem[\protect\astroncite{Hai {et~al.}}{2021}]{hai2021regular}
Hai Y, Guo Y, Cheng S, {et~al.}, 2021.
\newblock Regular position-oriented method for mesh smoothing.
\newblock {\em Acta Mechanica Solida Sinica},  34:437-448.

\bibitem[\protect\astroncite{Han {et~al.}}{2022}]{c34}
Han X, Gao H, Pfaff T, {et~al.}, 2022.
\newblock Predicting physics in mesh-reduced space with temporal attention.
\newblock {\em arXiv preprint arXiv:220109113}, .

\bibitem[\protect\astroncite{Herrmann}{1976}]{c12}
Herrmann LR, 1976.
\newblock Laplacian-isoparametric grid generation scheme.
\newblock {\em Journal of the Engineering Mechanics Division},  102(5):749-756.

\bibitem[\protect\astroncite{Kingma and Ba}{2014}]{c40}
Kingma DP, Ba J, 2014.
\newblock Adam: A method for stochastic optimization.
\newblock {\em arXiv preprint arXiv:14126980}, .

\bibitem[\protect\astroncite{Kipf and Welling}{2016}]{c30}
Kipf TN, Welling M, 2016.
\newblock Semi-supervised classification with graph convolutional networks.
\newblock {\em arXiv preprint arXiv:160902907}, .

\bibitem[\protect\astroncite{Klingner and
  Shewchuk}{2007}]{klingner2007aggressive}
Klingner BM, Shewchuk JR, 2007.
\newblock Aggressive tetrahedral mesh improvement.
\newblock IMR, p.3-23.

\bibitem[\protect\astroncite{Knupp}{2001}]{c7}
Knupp PM, 2001.
\newblock Algebraic mesh quality metrics.
\newblock {\em SIAM journal on scientific computing},  23(1):193-218.

\bibitem[\protect\astroncite{LeCun {et~al.}}{2015}]{c24}
LeCun Y, Bengio Y, Hinton G, 2015.
\newblock Deep learning.
\newblock {\em nature},  521(7553):436-444.

\bibitem[\protect\astroncite{Lee and Schachter}{1980}]{c39}
Lee DT, Schachter BJ, 1980.
\newblock Two algorithms for constructing a delaunay triangulation.
\newblock {\em International Journal of Computer \& Information Sciences},
  9(3):219-242.

\bibitem[\protect\astroncite{Li {et~al.}}{2019}]{c36}
Li G, Muller M, Thabet A, {et~al.}, 2019.
\newblock Deepgcns: Can gcns go as deep as cnns?
\newblock Proceedings of the IEEE/CVF international conference on computer
  vision, p.9267-9276.

\bibitem[\protect\astroncite{Liaw {et~al.}}{2018}]{tune}
Liaw R, Liang E, Nishihara R, {et~al.}, 2018.
\newblock Tune: A research platform for distributed model selection and
  training.
\newblock {\em arXiv preprint arXiv:180705118}, .

\bibitem[\protect\astroncite{Lino {et~al.}}{2021}]{c32}
Lino M, Cantwell C, Bharath AA, {et~al.}, 2021.
\newblock Simulating continuum mechanics with multi-scale graph neural
  networks.
\newblock {\em arXiv preprint arXiv:210604900}, .

\bibitem[\protect\astroncite{Liu {et~al.}}{2009}]{liu2009centroidal}
Liu Y, Wang W, L{\'e}vy B, {et~al.}, 2009.
\newblock On centroidal voronoi tessellation—energy smoothness and fast
  computation.
\newblock {\em ACM Transactions on Graphics (ToG)},  28(4):1-17.

\bibitem[\protect\astroncite{Lloyd}{1982}]{c15}
Lloyd S, 1982.
\newblock Least squares quantization in pcm.
\newblock {\em IEEE transactions on information theory},  28(2):129-137.

\bibitem[\protect\astroncite{Lopes}{2023}]{torchjit}
Lopes NP, 2023.
\newblock Torchy: A tracing jit compiler for pytorch.
\newblock Proceedings of the 32nd ACM SIGPLAN International Conference on
  Compiler Construction, p.98-109.

\bibitem[\protect\astroncite{Pak and Kim}{2017}]{c25}
Pak M, Kim S, 2017.
\newblock A review of deep learning in image recognition.
\newblock 2017 4th international conference on computer applications and
  information processing technology (CAIPT), p.1-3.

\bibitem[\protect\astroncite{Pan {et~al.}}{2020}]{pan2020hlo}
Pan W, Lu X, Gong Y, {et~al.}, 2020.
\newblock Hlo: Half-kernel laplacian operator for surface smoothing.
\newblock {\em Computer-Aided Design},  121:102807.

\bibitem[\protect\astroncite{Papagiannopoulos {et~al.}}{2021}]{teach2mesh}
Papagiannopoulos A, Clausen P, Avellan F, 2021.
\newblock How to teach neural networks to mesh: Application on 2-d simplicial
  contours.
\newblock {\em Neural Networks},  136:152-179.

\bibitem[\protect\astroncite{Parthasarathy and Kodiyalam}{1991}]{c17}
Parthasarathy V, Kodiyalam S, 1991.
\newblock A constrained optimization approach to finite element mesh smoothing.
\newblock {\em Finite Elements in Analysis and Design},  9(4):309-320.

\bibitem[\protect\astroncite{Paszy{\'n}ski {et~al.}}{2021}]{paszynski2021deep}
Paszy{\'n}ski M, Grzeszczuk R, Pardo D, {et~al.}, 2021.
\newblock Deep learning driven self-adaptive hp finite element method.
\newblock International Conference on Computational Science, p.114-121.

\bibitem[\protect\astroncite{Peng {et~al.}}{2022}]{c35}
Peng W, Yuan Z, Wang J, 2022.
\newblock Attention-enhanced neural network models for turbulence simulation.
\newblock {\em Physics of Fluids},  34(2).

\bibitem[\protect\astroncite{Pfaff {et~al.}}{2020}]{c33}
Pfaff T, Fortunato M, Sanchez-Gonzalez A, {et~al.}, 2020.
\newblock Learning mesh-based simulation with graph networks.
\newblock {\em arXiv preprint arXiv:201003409}, .

\bibitem[\protect\astroncite{Prasad}{2018}]{c10}
Prasad T, 2018.
\newblock A comparative study of mesh smoothing methods with flipping in 2d and
  3d.
\newblock Rutgers University-Camden Graduate School.

\bibitem[\protect\astroncite{Reddy}{1976}]{c23}
Reddy DR, 1976.
\newblock Speech recognition by machine: A review.
\newblock {\em Proceedings of the IEEE},  64(4):501-531.

\bibitem[\protect\astroncite{Samstag {et~al.}}{2016}]{c4}
Samstag RW, Ducoste JJ, Griborio A, {et~al.}, 2016.
\newblock Cfd for wastewater treatment: an overview.
\newblock {\em Water Science and Technology},  74(3):549-563.

\bibitem[\protect\astroncite{Sharp and Crane}{2020}]{sharp2020laplacian}
Sharp N, Crane K, 2020.
\newblock A laplacian for nonmanifold triangle meshes.
\newblock Computer Graphics Forum,  39(5):69-80.

\bibitem[\protect\astroncite{Song {et~al.}}{2022}]{c31}
Song W, Zhang M, Wallwork JG, {et~al.}, 2022.
\newblock M2n: mesh movement networks for pde solvers.
\newblock {\em Advances in Neural Information Processing Systems},
  35:7199-7210.

\bibitem[\protect\astroncite{Spalart and Venkatakrishnan}{2016}]{c1}
Spalart PR, Venkatakrishnan V, 2016.
\newblock On the role and challenges of cfd in the aerospace industry.
\newblock {\em The Aeronautical Journal},  120(1223):209-232.

\bibitem[\protect\astroncite{Tingfan {et~al.}}{2022}]{tingfan2022mesh}
Tingfan W, Xuejun L, Wei A, {et~al.}, 2022.
\newblock A mesh optimization method using machine learning technique and
  variational mesh adaptation.
\newblock {\em Chinese Journal of Aeronautics},  35(3):27-41.

\bibitem[\protect\astroncite{Ulyanov {et~al.}}{2016}]{c38}
Ulyanov D, Vedaldi A, Lempitsky V, 2016.
\newblock Instance normalization: The missing ingredient for fast stylization.
\newblock {\em arXiv preprint arXiv:160708022}, .

\bibitem[\protect\astroncite{Vartziotis and
  Papadrakakis}{2017}]{vartziotisImprovedGETMeAdaptive2017}
Vartziotis D, Papadrakakis M, 2017.
\newblock Improved {GETMe} by adaptive mesh smoothing.
\newblock {\em Computer Assisted Methods in Engineering and Science},
  20(1):55-71 (Number: 1).
\newblock  \\ https://cames.ippt.pan.pl/index.php/cames/article/view/80

\bibitem[\protect\astroncite{Vartziotis and Wipper}{2018}]{vartziotis2018getme}
Vartziotis D, Wipper J, 2018.
\newblock The getme mesh smoothing framework: A geometric way to quality finite
  element meshes.
\newblock CRC Press.

\bibitem[\protect\astroncite{Vartziotis
  {et~al.}}{2008}]{vartziotisMeshSmoothingUsing2008}
Vartziotis D, Athanasiadis T, Goudas I, {et~al.}, 2008.
\newblock Mesh smoothing using the {Geometric} {Element} {Transformation}
  {Method}.
\newblock {\em Computer Methods in Applied Mechanics and Engineering},
  197(45):3760-3767.
\newblock  \\
  https://www.sciencedirect.com/science/article/pii/S0045782508000996 \\
  https://doi.org/10.1016/j.cma.2008.02.028

\bibitem[\protect\astroncite{Vollmer {et~al.}}{1999}]{c19}
Vollmer J, Mencl R, Mueller H, 1999.
\newblock Improved laplacian smoothing of noisy surface meshes.
\newblock Computer graphics forum,  18(3):131-138.

\bibitem[\protect\astroncite{Wallwork {et~al.}}{2022}]{wallwork2022e2n}
Wallwork JG, Lu J, Zhang M, {et~al.}, 2022.
\newblock E2n: error estimation networks for goal-oriented mesh adaptation.
\newblock {\em arXiv preprint arXiv:220711233}, .

\bibitem[\protect\astroncite{Wang {et~al.}}{2022}]{c28}
Wang Z, Chen X, Li T, {et~al.}, 2022.
\newblock Evaluating mesh quality with graph neural networks.
\newblock {\em Engineering with Computers},  38(5):4663-4673.

\bibitem[\protect\astroncite{Wu {et~al.}}{2020}]{c18}
Wu Z, Pan S, Chen F, {et~al.}, 2020.
\newblock A comprehensive survey on graph neural networks.
\newblock {\em IEEE transactions on neural networks and learning systems},
  32(1):4-24.

\bibitem[\protect\astroncite{Xu {et~al.}}{2018}]{c22}
Xu K, Gao X, Chen G, 2018.
\newblock Hexahedral mesh quality improvement via edge-angle optimization.
\newblock {\em Computers \& Graphics},  70:17-27.

\bibitem[\protect\astroncite{Zhang {et~al.}}{2009}]{zhang2009surface}
Zhang Y, Bajaj C, Xu G, 2009.
\newblock Surface smoothing and quality improvement of quadrilateral/hexahedral
  meshes with geometric flow.
\newblock {\em Communications in Numerical Methods in Engineering},
  25(1):1-18.

\bibitem[\protect\astroncite{Zhang {et~al.}}{2020}]{meshingnet}
Zhang Z, Wang Y, Jimack PK, {et~al.}, 2020.
\newblock Meshingnet: A new mesh generation method based on deep learning.
\newblock International Conference on Computational Science, p.186-198.

\bibitem[\protect\astroncite{Zhang {et~al.}}{2021}]{meshingnet3d}
Zhang Z, Jimack PK, Wang H, 2021.
\newblock Meshingnet3d: Efficient generation of adapted tetrahedral meshes for
  computational mechanics.
\newblock {\em Advances in Engineering Software},  157:103021.

\bibitem[\protect\astroncite{Zhou and Shimada}{2000}]{c13}
Zhou T, Shimada K, 2000.
\newblock An angle-based approach to two-dimensional mesh smoothing.
\newblock {\em IMR},  2000:373-384.

\end{thebibliography}

\clearpage
\appendix
\section{Appendix}
\subsection{Examples of the mesh dataset}\label{appendix_data}
\begin{figure}[H]
    \centering
    \subfloat{\includegraphics[width=.45\columnwidth]{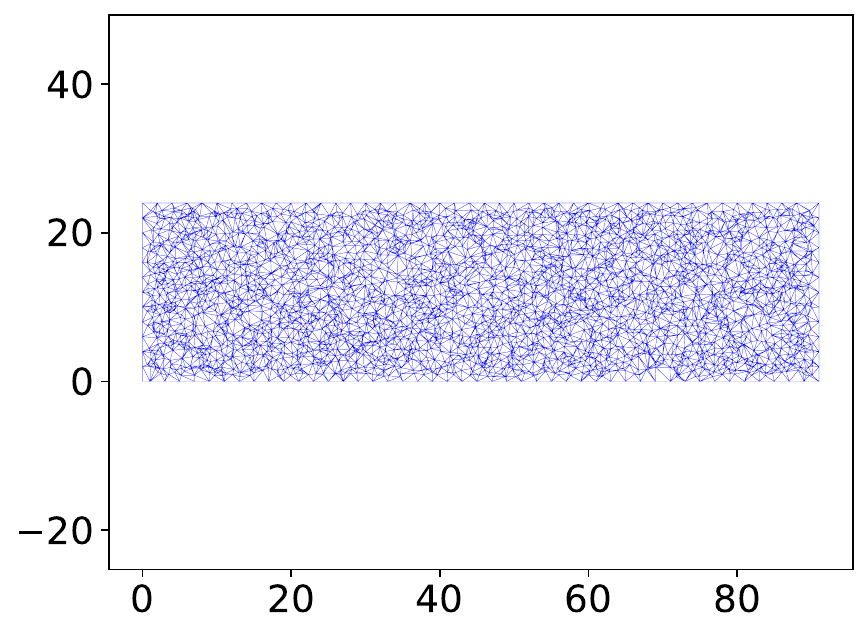}}\hspace{10pt}
    \subfloat{\includegraphics[width=.45\columnwidth]{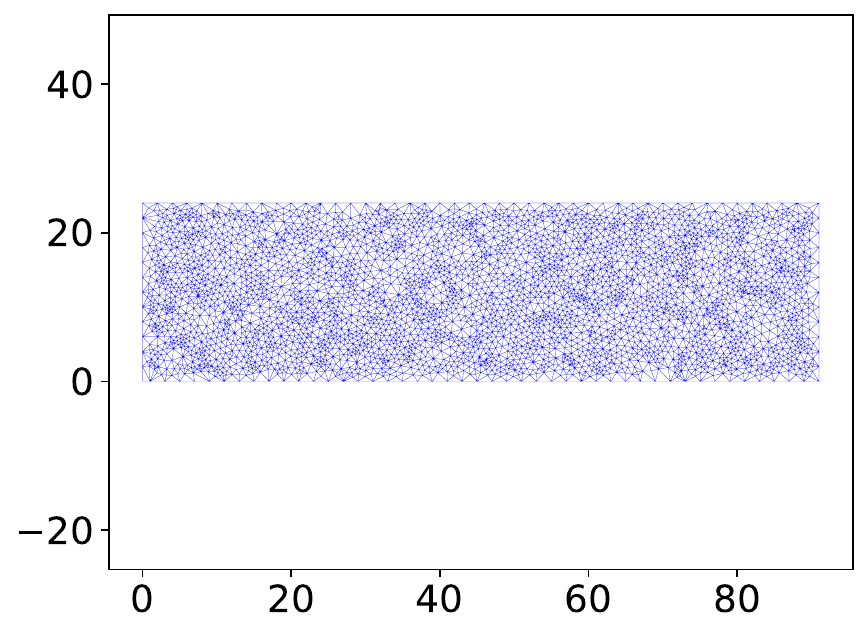}}\\
    \subfloat{\includegraphics[width=.45\columnwidth]{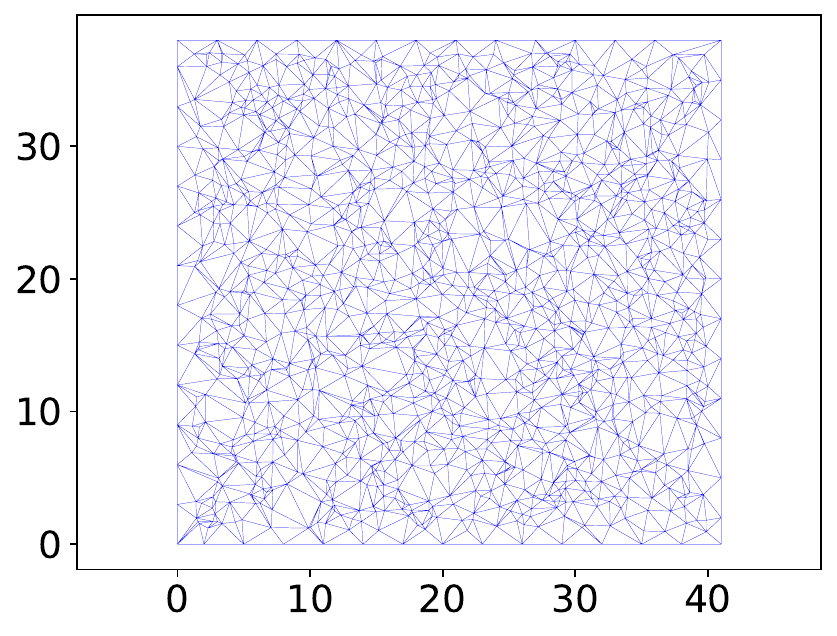}}\hspace{10pt}
    \subfloat{\includegraphics[width=.45\columnwidth]{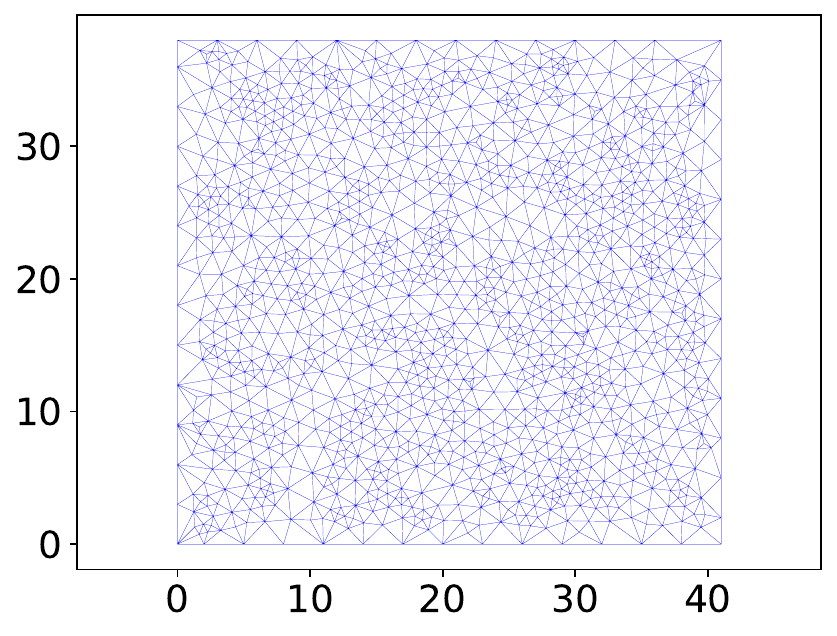}}\\
    \subfloat{\includegraphics[width=.45\columnwidth]{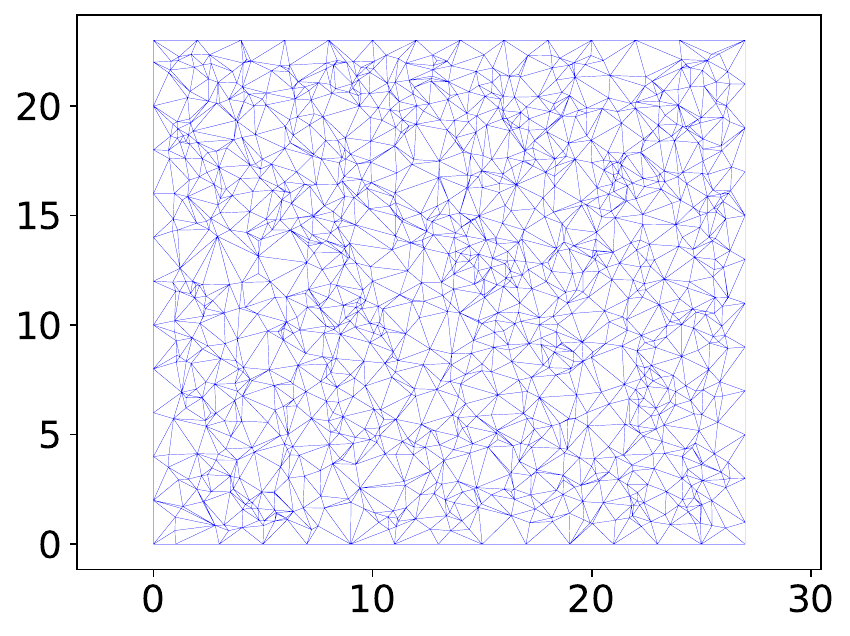}}\hspace{10pt}
    \subfloat{\includegraphics[width=.45\columnwidth]{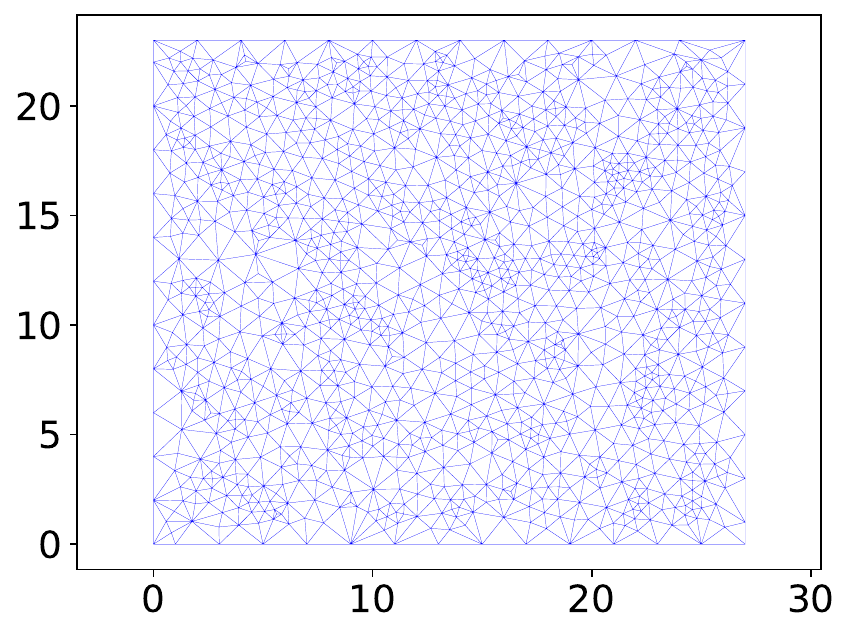}}\\
    \subfloat{\includegraphics[width=.45\columnwidth]{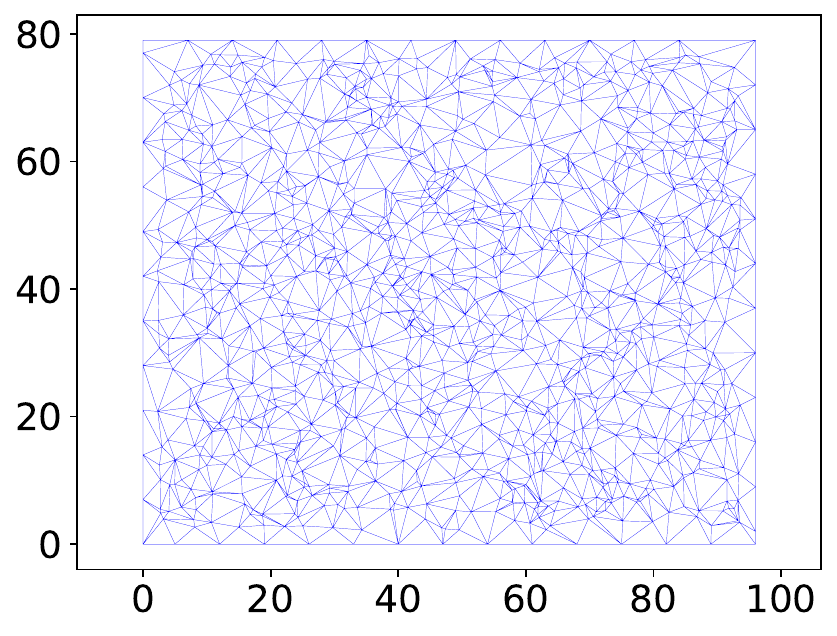}}\hspace{10pt}
    \subfloat{\includegraphics[width=.45\columnwidth]{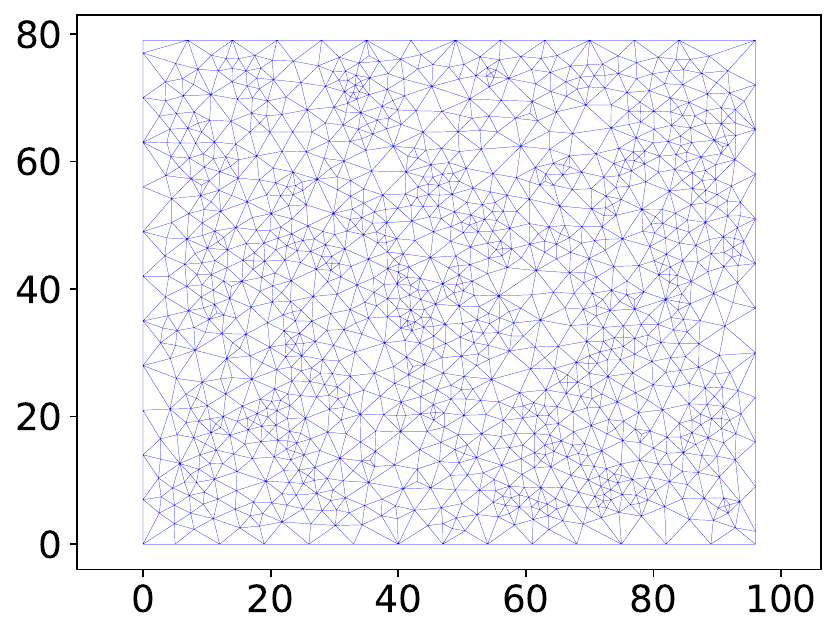}}\\
    \caption{Examples of meshes in the dataset. The left column represents the model input for GMSNet and NN-Smoothing training, while the column on the right represents the labeled mesh data to train NN-Smoothing model.}\label{more_data}
\end{figure}
\newpage
Here, we present four samples from the datasets used to train the NN-Smoothing and GMSNet models. As depicted in Figure \ref{more_data}, the meshes are only different  in  mesh size and element density.

\end{document}